\pdfoutput=1
\documentclass[11pt]{article}

\usepackage[final]{acl}

\usepackage{amssymb}
\usepackage{latexsym}

\usepackage[T1]{fontenc}
\usepackage[utf8]{inputenc}

\usepackage{microtype}

\usepackage{inconsolata}

\usepackage{graphicx}
\usepackage{float}

\usepackage{hyperref}
\usepackage{url}
\usepackage{booktabs}
\usepackage{multirow}
\usepackage[normalem]{ulem}
\usepackage{subcaption}
\usepackage[justification=centering]{caption}
\usepackage{adjustbox}
\usepackage{subcaption}
\usepackage{makecell}
\useunder{\uline}{\ul}{}
\usepackage{threeparttable}
\usepackage{siunitx}
\usepackage{tcolorbox}
\tcbuselibrary{breakable}
\usepackage[english]{babel}
\usepackage{kotex}
\usepackage{wrapfig}
\usepackage{mdframed}
\usepackage{array}
\usepackage{enumitem}
\usepackage[T1]{fontenc}
\usepackage{tabularx}
\usepackage[table,xcdraw]{xcolor}
\usepackage{colortbl}
\usepackage{bbm}
\usepackage{amsmath}
\usepackage{adjustbox}

\title{Uncovering Factor-Level Preference to Improve Human-Model Alignment}

\author{
Juhyun Oh\thanks{\,Equal contribution.}$^{\,1}$ \quad
Eunsu Kim\footnotemark[1]$^{\,1}$ \quad
Jiseon Kim$^{1}$ \quad
Wenda Xu$^{2}$ \quad
Inha Cha$^{3}$ \\
\textbf{William Yang Wang$^{2}$ \quad Alice Oh$^{1}$} \\
\\
$^{1}$KAIST \quad
$^{2}$University of California, Santa Barbara \quad
$^{3}$Georgia Institute of Technology \\
\texttt{\{411juhyun, kes0317\}@kaist.ac.kr}
}
\begin{document}
\maketitle
\begin{abstract}
Large language models (LLMs) often exhibit tendencies that diverge from human preferences, such as favoring certain writing styles or producing overly verbose outputs. 
While crucial for improvement, identifying the factors driving these misalignments remains challenging due to existing evaluation methods' reliance on coarse-grained comparisons and lack of explainability.
% However, current methods for evaluating preference alignment often lack explainability, relying on coarse-grained comparisons.
To address this, we introduce PROFILE, an automated framework to uncover and measure factor-level preference alignment of humans and LLMs.
Using PROFILE, we analyze preference alignment across three key tasks: summarization, instruction-following, and document-based QA. We find a significant discrepancy: while LLMs show poor factor-level alignment with human preferences when generating texts, they demonstrate strong alignment in discrimination tasks. 
We demonstrate how leveraging the identified generation-discrimination gap can be used to improve LLM alignment through multiple approaches, including fine-tuning with self-guidance.
Our work highlights the value of factor-level analysis for identifying hidden misalignments and provides a practical framework for improving LLM-human preference alignment.
% Our study offers a deeper understanding of human-model preference misalignment and provides a pathway towards developing LLMs more genuinely aligned with nuanced human values.
% Human preference for generated text is multifaceted, with the importance of factors like conciseness, helpfulness, and fluency varying by task. However, it remains unclear whether LLMs prioritize these factors in a human-aligned manner during generation. 
% We introduce PROFILE, an automated framework to quantify and compare factor-level preferences between humans and models. Using PROFILE, we find that models exhibit systematic misalignments with human preferences when generating text, notably a consistent bias towards longer outputs across diverse tasks, even when detrimental to task-specific quality. Interestingly, our analysis also reveals that these models demonstrate significantly better alignment with human factor preferences when evaluating text. Leveraging this insight, we propose an automatic method that uses a model's more human-aligned evaluation capabilities to guide its generation process, thereby reducing misalignment. PROFILE offers a deeper understanding of model generation priorities and provides a pathway towards developing LLMs more genuinely aligned with nuanced human values

\end{abstract}

\section{Introduction}
% \input{Text/1_Introduction_new}
% multi-faceted, why we have to look at factor-level ... 
Human preference for text is inherently multifaceted, influenced by an interplay of factors such as fluency, helpfulness, and conciseness. The relative importance of these factors is not static; it shifts depending on the specific task and context. For instance, a desirable summary should be concise and to the point, while creative writing might prioritize novelty and an engaging narrative. 
As large language models (LLMs) generate increasingly human-like text, a critical question arises: 
do LLMs prioritize these quality factors in ways that align with human expectations?
% do these models truly capture the nuance of these varied human expectations?

% , particularly in how they prioritize underlying quality factors—what we term factor-level preference—when generating responses?

\begin{figure}[!t]
    \centering
    \includegraphics[width=0.95\columnwidth]{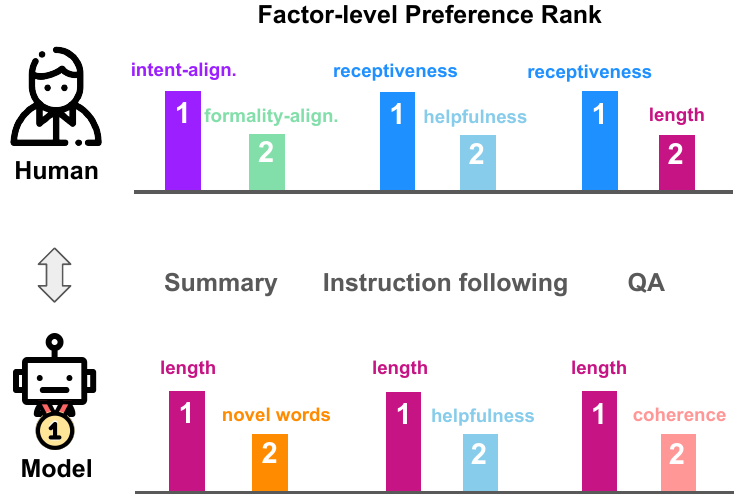}
    % \caption{\revise{so long! }LLMs often fall short of expectations, but pinpointing \textit{why} is challenging. PROFILE quantifies the influence of various factors on preferences, revealing the underlying causes of human-model misalignment.}
    \caption{PROFILE uncovers that models exhibit misalignments with human preferences when generating texts. While humans prioritize different quality factors for different tasks, models show consistent bias towards longer output.}
    \label{fig:main}
    \vspace{-3mm}
\end{figure}
\begin{figure*}[!t]
    \centering
    \includegraphics[width=0.9\textwidth]{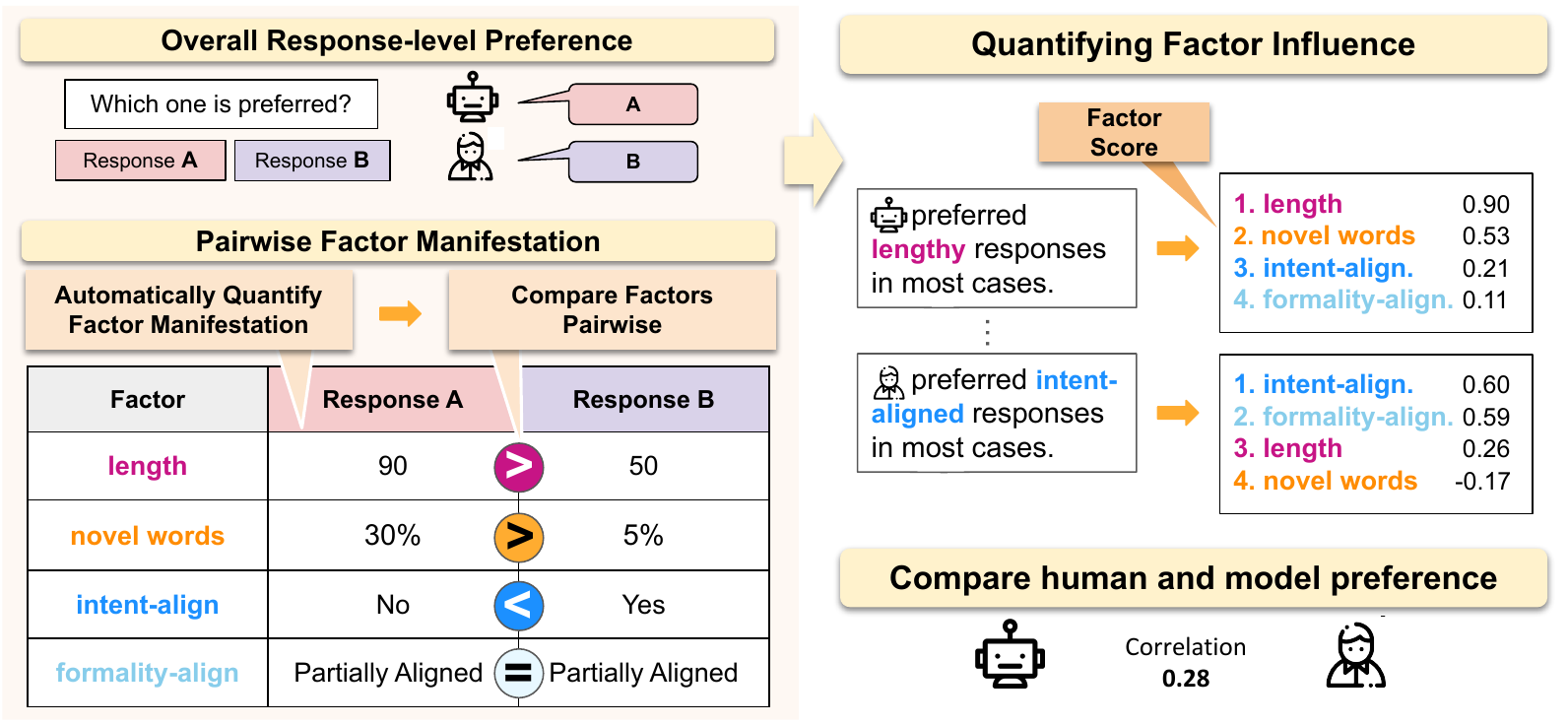}
    \caption{\textbf{An overview of PROFILE pipeline}: (1) Extracting overall Response-level Preference, (2) Comparing factor manifestation in a pairwise manner, (3) Quantifying Factor Influence, and (4) Comparing human and model preference at the factor-level.}
    \label{fig:pipeline}
    % \vspace{-3mm}
\end{figure*}
% do these models inherently prioritize these quality factors in a way that aligns with human expectations when they generate responses?

% 어떻게 이걸 소개하고 연결해야할지 모르겠다 ㅜㅠ
% This question is particularly relevant given existing research highlighting a discrepancy between LLM's internal encoding (parametric knowledge) and external behavior~\cite{orgad2024llms}. Notably, recent studies~\cite{west2023generative, oh-etal-2024-generative} suggest that a model's ability to generate responses and its ability to discriminate between them are not necessarily aligned. We extend this line of inquiry to investigate whether this discrepancy also manifests at the factor-level of preference alignment. Specifically, do models prioritize individual quality factors consistently with human judgment during generation, and does this prioritization differ when they are asked to discriminate between outputs?

This question becomes particularly pressing given recent findings highlighting discrepancies between LLMs' generation and discrimination abilities~\cite{west2023generative, oh-etal-2024-generative}. We extend this line of inquiry to investigate whether such discrepancies also manifest at the factor level of preference alignment: do models prioritize individual quality factors consistently with human judgment during \textit{generation}, and does this prioritization differ when \textit{discriminating} between responses?

Despite significant advances in preference alignment~\citep{ouyang2022training, rafailov2024direct, song2024preference}, existing approaches take a coarse-grained view, measuring overall preferences while overlooking the underlying factors that drive them. Recent work has begun examining finer-grained preference aspects~\citep{hu-etal-2023-decipherpref, kirk2024prism, scheurer2023training}, yet systematic factor-level comparisons between human and model priorities remain limited. Moreover, existing research has predominantly focused on generation settings, leaving open whether models exhibit consistent factor prioritization across different task settings.

% \revise{Previous attempts to understand and improve preference alignment~\citep{ouyang2022training,  rafailov2024direct, song2024preference} have largely taken a coarse-grained view, measuring which model is preferred overall but offering little insight into the underlying factors that drive these preferences. Some recent studies move toward finer-grained analyses of human preferences~\citep{hu-etal-2023-decipherpref, kirk2024prism, scheurer2023training}, yet systematic comparisons between human and model factor-level preferences remain limited. Moreover, existing work has predominantly examined alignment in the \textit{generation} setting, leaving open whether models exhibit the same prioritization of quality factors when tasked with \textit{discrimination} (i.e., choosing among multiple options). Prior studies suggest that models with similar parametric knowledge can behave differently across these settings~\cite{west2023generative, oh-etal-2024-generative}, motivating a closer factor-level investigation.}

To address these gaps, we introduce PROFILE, an automated framework designed to decompose and quantify how individual factors (\emph{e.g.}, fluency, helpfulness) contribute to overall preference decisions. 
PROFILE quantifies each factor's contribution as a \textbf{factor score}. By comparing how factor scores of responses correlate with overall preference decisions, PROFILE captures the factor-level preferences of humans and models.
% PROFILE measures factor-level alignment by comparing how differences in specific factors between responses correlate with the overall preference decisions of humans and models.
This enables systematic comparison between human and model priorities across both generation and discrimination settings.
% PROFILE allows us to systematically compare what humans value against what models do, moving beyond surface-level quality scores to uncover deeper, factor-level alignments and misalignments. 
Using this framework, we investigate three key research questions:
\begin{enumerate}
    \item To what extent do LLMs, during generation, exhibit factor-level preference alignment with human expectations across various tasks?
    \item How does this factor-level alignment compare when the same models perform discrimination tasks (\emph{i.e.}, distinguishing between good and bad responses) versus generation?
    \item Can insights from observed alignment differences between these settings be leveraged to improve the less aligned setting?
\end{enumerate}

We conduct comprehensive experiments across three preference alignment tasks---summarization, instruction-following, and document-based QA---evaluating eight prominent LLMs.  Our analysis reveals systematic misalignments: models often do not prioritize quality factors in line with human expectations during generation. For instance, models frequently exhibit a strong preference for length regardless of the task, whereas human preferences for factors such as conciseness or detail vary contextually (Figure~\ref{fig:main}). 

Interestingly, we observe that these same LLMs demonstrate notably better factor-level alignment during discrimination tasks, specifically in evaluation settings where the model selects which of two outputs is better. This disparity between generation and discrimination alignment presents an opportunity, and we demonstrate ways to leverage the stronger alignment in discrimination to enhance the factor-level preference alignment during generation. Our work underscores the importance of factor-level analysis for a deeper understanding of LLM alignment and offers a pathway toward more genuinely human-aligned generative models.
% \input{Figure/pipeline}

% \section{Related Work}
% \input{Text/2_Related_Work}

% \section{Problem Definition}
% \input{Text/3_Experiments_new}
% \section{\textbf{PROFILE}: Measuring preference alignment by \textbf{PRO}bing \textbf{F}actors of \textbf{I}nf\textbf{L}uence for \textbf{E}xplainability}
\section{PROFILE: Framework for Analyzing Human–Model Alignment}    
\label{sec:profile}

We introduce PROFILE (Probing Factors of Influence for Explainability), a framework that provides a systematic way to decompose overall response level preferences into their underlying factors. PROFILE makes explicit the notion of \textit{factor-level preference}---the extent to which specific factors such as fluency, factual accuracy, or conciseness influence which response is preferred. By uncovering these factor-level preferences, PROFILE enables a more interpretable analysis of human and model choices and provides a principled basis for comparing and improving their alignment.

PROFILE centers on the computation of \textbf{factor scores}, which quantify the influence of individual factors on overall preferences.
To compute these scores, PROFILE proceeds in three steps:
(1) we define overall response-level preferences for humans and models (\S~\ref{sec:method_overall_pref});
(2) we quantify the manifestation of each factor in every response based on our factor taxonomy (\S\ref{sec:factor_analysis});
(3) we compare these pairwise factor manifestations with preference labels to measure concordance, thereby deriving factor scores that reveal the influence of each factor and the alignment of model preferences with human values (\S\ref{sec:uncovering_pref} \& \S\ref{sec:comparison}).
Figure~\ref{fig:pipeline} illustrates this overall process, which is detailed below:

\begin{table*}[ht]
\centering
\small
\begin{tabularx}{\textwidth}{lXl}
\toprule
\rowcolor[rgb]{0.85,0.85,0.85}
\textbf{Factor} & \textbf{Description} & \textbf{Tasks} \\
\midrule
Receptiveness & Whether the core question of the input has been answered. & \textit{I}, \textit{Q} \\
Off Focus & The ratio of atomic facts that are not related to the main focus of the input. & \textit{S}, \textit{I}, \textit{Q} \\
Intent Align. & Whether the intent of the source and output is the same. & \textit{S} \\
Hallucination & The ratio of atomic facts that are incorrect compared to the original source. & \textit{S}, \textit{I}, \textit{Q} \\
Source Coverage & The ratio of atomic facts in the source that appear in the output. & \textit{S} \\
Formality Align. & Whether the formality of the source and output is the same. & \textit{S} \\
Novel Words & The ratio of words in the output that are not used in the source. & \textit{S} \\
Length & The number of words used in the output. & \textit{S}, \textit{I}, \textit{Q} \\
Fluency & The quality of individual sentences of the output. & \textit{S}, \textit{I}, \textit{Q} \\
Number Of Facts & The number of atomic facts in the output. & \textit{S}, \textit{I}, \textit{Q} \\
Helpfulness & The ratio of facts that provide additional helpful information. & \textit{I}, \textit{Q} \\
Misinformation & The ratio of facts in the output that include potentially incorrect or misleading information. & \textit{I}, \textit{Q} \\
Coherence & Whether all the sentences of the output form a coherent body. & \textit{S}, \textit{I}, \textit{Q} \\
\bottomrule
\end{tabularx}
\caption{The full taxonomy factors, definitions, and associated tasks (S: Summarization, I: Instruction-following, Q: DocumentQA).}
\label{tab:taxonomy}
\end{table*}

\subsection{Quantifying Factor Influence}
\label{sec:uncovering_pref}

To quantify the influence of a given factor $f$, we calculate its \textbf{factor score}, $\tau(f)$, by analyzing the concordance between response-level preferences and factor-level manifestations across all pairs of responses $\{r_i, r_j\}$. We use $\tau_{14}$, a variation of Kendall's correlation well-suited for handling ties~\citep{machacek-bojar-2014-results}, defined as:
\begin{equation} 
\label{eq:factorscore} 
\tau(f) = \frac{|C_f| - |D_f|}{|C_f| + |D_f| + |T_f|}
\end{equation}
Here, $C_f$ is the set of \textbf{concordant} pairs, where the overall preference aligns with the factor manifestation (\emph{i.e.}, the preferred response also has a stronger manifestation of the factor). $D_f$ is the set of \textbf{discordant} pairs where they do not align, and $T_f$ accounts for ties.

A positive score indicates a positive preference for the factor, a negative score indicates a negative preference, and a score near zero implies minimal influence. The magnitude reflects the strength of this influence. To compute this score, we must first define its two key components: the overall response-level preference ($Pref$), and the pairwise factor manifestation ($M_f$).

\subsection{Measuring Overall Response-level Preference}
\label{sec:method_overall_pref}
The overall preference function $Pref(r_i, r_j)$ captures which response is considered better in a pair. We define it as:
\[ Pref(r_i, r_j) = \operatorname{sign}\!\big(U(r_i) - U(r_j)\big), \]
where $U(r)$ denotes the value assigned to a response $r$ by the agent. This yields $1$ if $r_i$ is preferred, $-1$ if $r_j$ is preferred, and $0$ for a tie. This value function is obtained as follows for humans and models.

\textbf{Human Preference.} For humans, the value $U(r)$ is derived directly from pairwise annotations where labelers select the preferred response, thus determining the sign of the preference. In our study, we leverage existing datasets with human preference labels to obtain these values.

\textbf{Model Preference in Generation.} A model's generation preference is traditionally defined using log likelihood ($P(x) = \sum_{i=1}^{n} \log P(x_i | x_{<i})$). While this is a direct measure, it presents practical challenges: manipulating logits to obtain distinctive outputs can be difficult, and log probabilities are often inaccessible for closed models. 

To overcome these issues, we use \textit{score-based prompting} as a proxy measure. In this approach, we instruct the LLM to generate a response conditioned on achieving a target quality score from 1 to 5. The target score itself serves as the value for the generated response, $U(r)$. For instance, if response $r_i$ was generated with a target score of 4 and $r_j$ with a score of 3, we define that $U(r_i) > U(r_j)$, and thus the model ``prefers'' $r_i$ in this generation context. This approach is inspired by methods used in constructing training datasets for LLM-as-a-judge~\citep{kim2023prometheus}, reflecting real-world applications where models are conditioned on specific quality targets.

To validate that this proxy effectively approximates the models’ intrinsic preferences, we conducted an experiment using 100 samples from summarization tasks. Specifically, we prompted open-source models (Llama-3.1-70B and Mixtral) to generate distinct summaries for target scores ranging from 1 to 5. We then computed the log probability of each generated summary and observed a strong Pearson correlation with the target scores (Llam: 0.975; Mixtral: 0.82; see Figure~\ref{fig:method-valid} in the Appendix). These results suggest that our scoring mechanism serves as an effective proxy for the models' intrinsic generation preferences.
\subsection{Measuring Pairwise Factor Manifestation}
\label{sec:method_factor_manifest}
% \paragraph{Pairwise Factor Manifestation ($M_f$).}
% \label{sec:pairwise_comparison}
The factor manifestation function $M_f(r_i, r_j)$ determines which response in a pair exhibits a stronger presence of factor $f$. It is defined similarly to preference:
\[ M_f(r_i, r_j) = \operatorname{sign}\!\big(m_f(r_i) - m_f(r_j)\big), \]
where $m_f(r)$ is a scalar measurement of factor $f$ in response $r$. For example, if $m_{\mathsf{length}}(r)$ is the character count, $M_{\mathsf{length}}(r_i, r_j) = 1$ when $r_i$ is longer. The measurement $m_f(r)$ is derived from our factor taxonomy.

\paragraph{Taxonomy of Preference Factors.}
\label{sec:factor_analysis}
To provide a structured framework for analyzing preferences across diverse text generation tasks, we develop a unified taxonomy of fine-grained factors relevant to text quality. 
This taxonomy categorizes the factors influencing preference alignment between humans and LLMs across text generation tasks. 
Addressing the lack of a unified framework and inconsistent terminology in existing literature, we consolidate evaluation factors from diverse tasks, including summarization, instruction following, and question answering. For summarization-specific factors, we draw from \citet{fu2024gptscore, hu-etal-2023-decipherpref, zhong-etal-2022-towards, fabbri2021summeval}. For instruction-following and document-based question answering, we incorporate categories from \citet{glaese2022improving, ye2024flask, nakano2021webgpt}. The complete taxonomy is detailed in Table~\ref{tab:taxonomy}.

% To quantify the manifestation $m_f(r)$ for each factor, we use a combination of automated methods:
% (i) \textbf{Rule-based}: For objective factors like $\mathsf{Length}$.
% (ii) \textbf{UniEval-based}: For subjective factors like $\mathsf{Coherence}$~\citep{zhong-etal-2022-towards}.
% (iii) \textbf{LLM-based}: For nuanced factors like $\mathsf{Intent\:Alignment}$, using GPT-4o.
% Further implementation details are in Appendix~\ref{sec:appendix_profile}.

\paragraph{Quantifying Factor Manifestation.}
We use several approaches to automatically analyze the manifestation of our factors in responses:
(i) \textbf{Rule-based}: For straightforward, objective factors, we use deterministic algorithms. $\mathsf{Length}$ and $\mathsf{Novel\:Words}$ are extracted this way. 
(ii) \textbf{UniEval-based}: For inherently subjective factors ($\mathsf{Fluency}$ and $\mathsf{Coherence}$), we use the well-established UniEval metric~\citep{zhong-etal-2022-towards}. UniEval is a learned metric that provides scores of range 0-1 for various aspects of text quality. 
(iii) \textbf{LLM-based}: For factors that rely on objective criteria but require more nuanced judgment, we use GPT-4o with carefully designed prompts. This approach is further divided into ``response-based'' ($\mathsf{Intent\:Alignment}$ and $\mathsf{Formality\:Alignment}$) and ``atomic-fact-based'' (the remaining seven) extraction depending on the level of detail needed for each factor.
The specific details of the implementation of each method and validation of LLM-based extractions can be found in Appendix~\ref{sec:appendix_profile}.

\subsection{Comparing Human and Model Preferences}
\label{sec:comparison}

Finally, we measure the \textbf{factor-level preference alignment} between humans and models. With a factor score $\tau(f)$ computed for each agent, we create a ranked list of factors for both humans and models. We then quantify the alignment of these rankings using standard correlation coefficients: Spearman’s $\rho$, Kendall’s $\tau_b$, and Pearson’s $r$. This provides a clear metric for how well a model's factor priorities align with human values.

Together, these steps make PROFILE a structured framework for analyzing the drivers of preference judgments. By making factor-level influences explicit, PROFILE enables 1) more interpretable comparisons between humans and models and 2) provides a consistent basis for assessing their alignment across different tasks and factors.

% \revise{PROFILE enables}:

% \begin{enumerate}[itemsep=0.5pt, topsep=1pt] 
%     \item \textbf{Interpretation} of overall response quality at a fine-grained factor level. It automatically measures how much each factor (e.g., helpfulness, conciseness, length) contributes to overall preference.
%     \item \textbf{Comparison of alignment between human and model preferences.} PROFILE provides a quantitative score that serves as a metric for human-model alignment in factor level.
% \end{enumerate}

% By making these aspects explicit, PROFILE provides a structured way to diagnose and improve alignment between model behavior and human values.

\section{Uncovering Factor-Level Preference of LLMs}
% \begin{figure}[h]
%     \centering
%     \includegraphics[width=0.8\textwidth]{Figure/summary_fig.pdf}
%     \caption{}
%     \label{fig:summary_main}
% \end{figure}

% \begin{figure}[h]
%     \centering
%     \includegraphics[width=0.8\textwidth]{Figure/shp2_fig.pdf}
%     \caption{}
%     \label{fig:shp_main}
% \end{figure}

% \begin{figure}[h]
%     \centering
%     \includegraphics[width=0.8\textwidth]{Figure/webgpt_fig.pdf}
%     \caption{}
%     \label{fig:web_main}
% \end{figure}
    \vspace{0.5cm}
\begin{figure*}[tp]
% \begin{figure*}[ht]
    \centering
    % Subfigure 1
    \begin{subfigure}[b]{\textwidth}
        \centering
        \includegraphics[width=\textwidth]{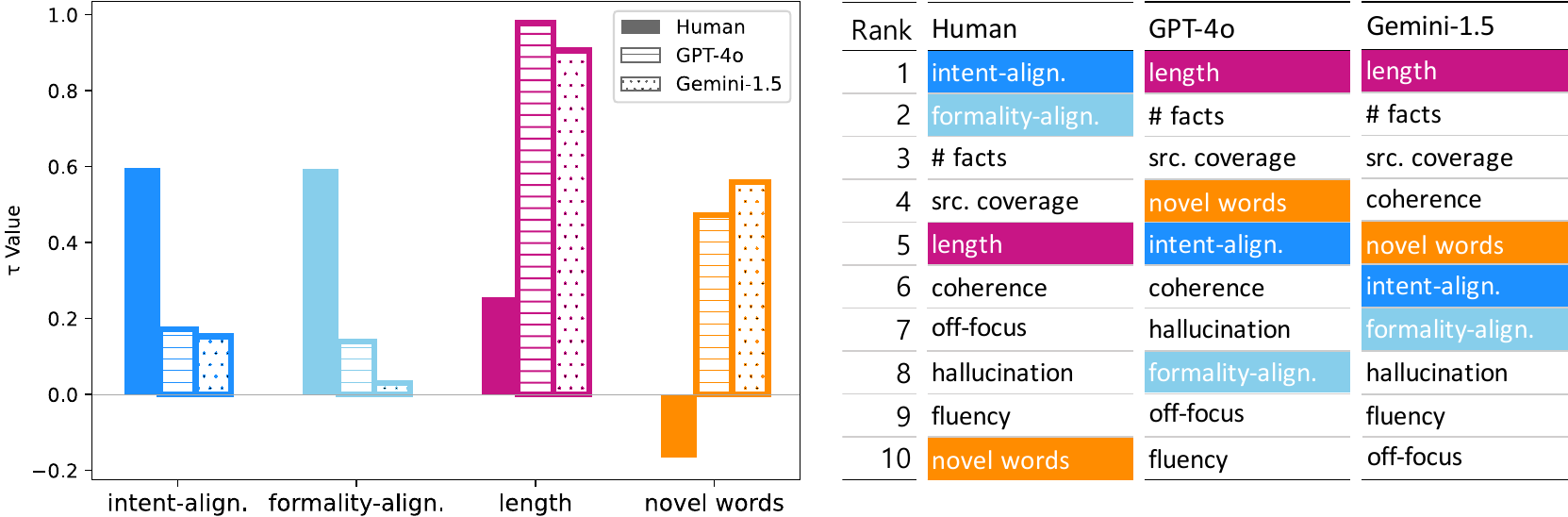}
        \caption{Summarization}
        \label{fig:summary_main}
    \end{subfigure}
    
    % \vspace{0.5cm} % Adjust the space between subfigures as needed

    % Subfigure 2
    \begin{subfigure}[b]{\textwidth}
        \centering
        \includegraphics[width=\textwidth]{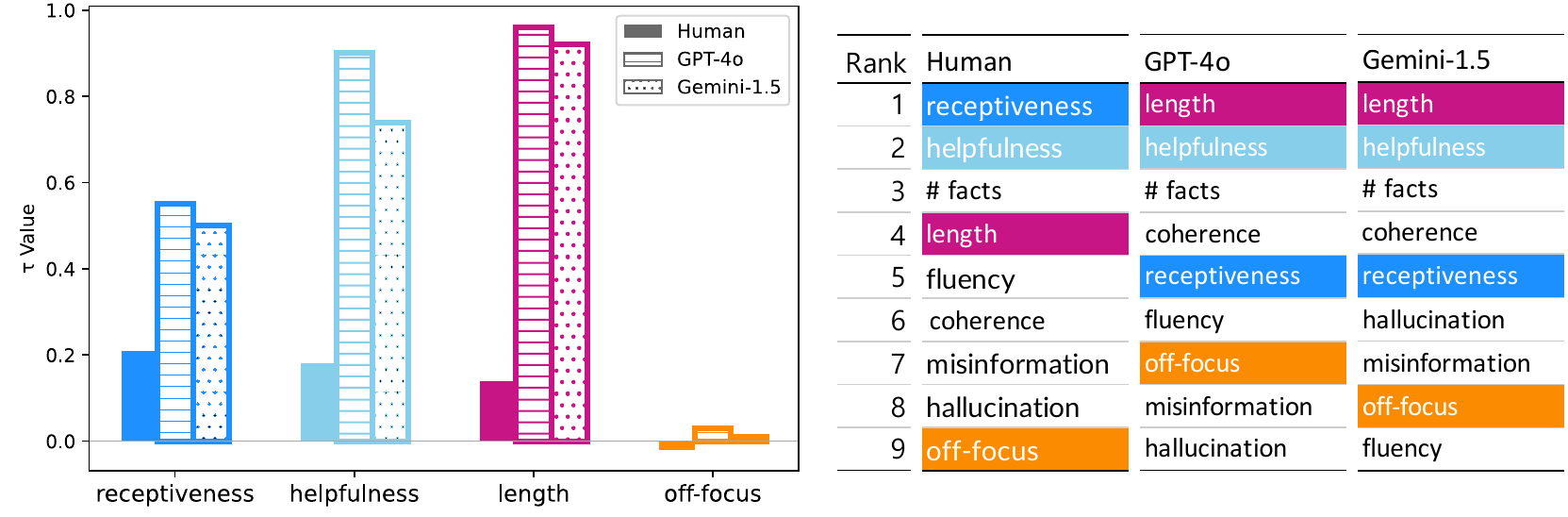}
        \caption{Instruction-following}
        \label{fig:shp_main}
    \end{subfigure}

        % Subfigure 2
    \begin{subfigure}[b]{\textwidth}
        \centering
        \includegraphics[width=\textwidth]{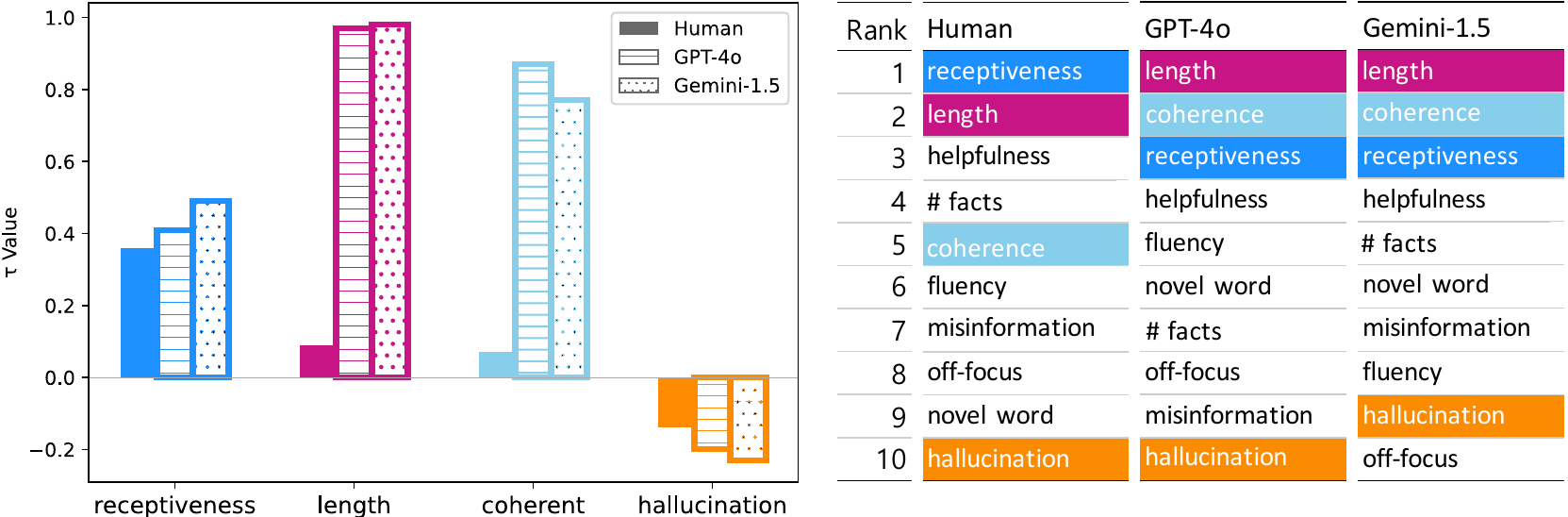}
        \caption{Document QA}
        \label{fig:web_main}
    \end{subfigure}
    
 % Adjust the space between subfigures as needed

    \caption{\textbf{PROFILE uncovers the factor-level preferences of humans and models.} Figure illustrates the comparison of factor-level preference alignment between humans, GPT-4o, and Gemini-1.5 in generation across three tasks: (a) Summarization, (b) Instruction-following, and (c) Document QA task. The left bar graphs display \textit{factor scores} ($\tau_{14}$) for selected factors. The right tables show the rankings of all factors for each task. Notably, both models consistently rank `length' as the top factor across tasks, while human preferences vary by task.}
    \label{fig:combined_main}
    \vspace{-30mm}
\end{figure*}

% \subsection{Experimental Overview: Analyzing Preferences in Generation and Evaluation}
\label{sec:pre-exp}

In this section, we analyze how models and humans differ in their factor-level preferences during text generation. Using human preference datasets across summarization, instruction-following, and QA tasks, we apply PROFILE to model-generated responses and compare the relative importance of each factor with human judgments.
\subsection{Experimental Setting}
\label{sec:exp-setting}
% \begin{table}[htbp]
%     \centering
%     \small
%     \begin{tabular}{@{}l|ccc@{}}
%         \toprule
%         & \multicolumn{3}{c}{\textbf{Generation}} \\ \midrule
%         & \textit{$\tau$} & \textit{$\rho$} & \textit{r} \\ \midrule
%         Mixtral       & 0.200 & 0.297 & 0.069 \\
%         Tulu 2.5 + PPO (13B RM) & -0.156 & -0.164 & -0.189 \\
%         Tulu 2.5 + PPO (70B RM) & 0.111 & 0.200 & -0.015 \\
%         LLaMA 3.1 70B & 0.111 & 0.248 & 0.213 \\
%         Gemini 1.5     & \textbf{0.289} & \textbf{0.394} & \textbf{0.171} \\
%         GPT-4o         & 0.156 & 0.297 & 0.155 \\ 
%         \bottomrule
%     \end{tabular}
%     \caption{Factor-level preference alignment (\textit{$\tau$}, \textit{$\rho$}, \textit{r}) between model and human in the generation setting for the summarization task.}
%     \label{tab:generation_correlations}
% \end{table}

\begin{table}[htbp]
    \centering
    \small
    \begin{tabular}{@{}l|ccc@{}}
        \toprule
        & \textit{$\tau$} & \textit{$\rho$} & \textit{r} \\ \midrule
        Mixtral       & 0.200 & 0.297 & 0.069 \\
        Tulu 2.5 + PPO (13B RM) & -0.156 & -0.164 & -0.189 \\
        Tulu 2.5 + PPO (70B RM) & 0.111 & 0.200 & -0.015 \\
        LLaMA 3.1 70B & 0.111 & 0.248 & 0.213 \\
        Gemini 1.5     & \textbf{0.289} & \textbf{0.394} & \textbf{0.171} \\
        GPT-4o         & 0.156 & 0.297 & 0.155 \\ 
        \bottomrule
    \end{tabular}
    \caption{Factor-level preference alignment (\textit{$\tau$}, \textit{$\rho$}, \textit{r}) between model and human in the generation setting for the summarization task.}
    \label{tab:generation_correlations}
    \vspace{-10pt}
\end{table}

\paragraph{Tasks.} We use human preference alignment data publicly available. Among them we choose: (i) Reddit TL;DR~\citep{stiennon2020learning}, which includes human ratings of summaries across multiple evaluation dimensions; (ii) StanfordHumanPreference-2 (SHP-2)~\citep{pmlr-v162-ethayarajh22a}, focusing on human preferences over responses in the ``\texttt{reddit/askacademia}'' domain; and (iii) OpenAI WebGPT~\citep{nakano2021webgpt}, which compares model-generated answers on the ELI5 subreddit based on factual accuracy and usefulness\thinspace\footnote{Our framework can also be applied to other tasks. 
We provide guidelines for applying it to different tasks, with an example of a mathematical reasoning task in the Appendix~\ref{appendix-apply-to-math}.}. We refer to the tasks for each dataset as summarization, instruction-following, and document-based QA tasks in this paper. We exclude pairs with human Tie ratings in all three datasets, as our analysis focuses on cases with clear preference distinctions.

\paragraph{Models.}
For our experiments, we utilize both open-source and proprietary LLMs. Open-source models include LLaMA 3.1 70B~\citep{dubey2024llama}, Mixtral 8x7B Instruct v0.1~\citep{jiang2024mixtral}, and three TÜLU v2.5 models ~\citep{ivison2024unpacking} (TÜLU v2.5 + PPO 13B (13B RM), TÜLU v2.5 + PPO 13B (70B RM), and TÜLU v2.5 + DPO 13B). Proprietary models include Gemini 1.5 Flash~\citep{reid2024gemini}, GPT-4o~\citep{openai2024hello_gpt4o}, and GPT-3.5. 
From here on, we refer to Gemini 1.5 Flash as Gemini 1.5, Mixtral 8x7B Instruct v0.1 as Mixtral, TÜLU v2.5 models as Tulu 2.5 + \{alignment training strategy\}.
Detailed descriptions of the datasets and models can be found in Appendix~\ref{sec:appendix_task_model}.

\paragraph{Experimental Setup.}
For each task, models generate a response that would receive a score of 1-5. The specific prompts we used can be found in Appendix~\ref{sec:appendix_prompts}.
% \revise{For each task, we explore two settings: (i) Generation, where models generate responses that would receive a score of 1-5 for a given task, and (ii) Evaluation, where models select the better of two provided responses, which are taken from the datasets. 
% See Appendix~\ref{sec:appendix_prompts} for prompts. }
% In both settings, we use PROFILE to extract factor scores and their factor rankings and measure the correlation with human judgments (factor-level preference alignment).
% In addition to factor-level analysis, we assess overall pairwise response agreement between humans and models. 
% \revise{For evaluation, we report the percentage of models' agreement with existing human labels by measuring how often it aligns with human judges' selections of the better response.}
Additionally, we find that responses generated with score 5 strongly align with those from direct, unconstrained generation (see Table~\ref{tab:response_score_similarity}), suggesting the generalizability of our experimental setting.
% For generation, we collect human evaluations via Amazon Mechanical Turk (AMT) for the summarization task. We assess whether the response that the model assigned a higher score to is also considered a better response by human judges. We report the percentage of agreement between the model's scores and human preferences.
\subsection{Factor-level Alignment in Generations}
\label{sec:gen_alignment}
PROFILE enables fine-grained analysis of preference alignment by breaking down overall judgments into interpretable factor-level scores. This allows us to identify not only how models and humans differ in ranking specific factors (Figure~\ref{fig:combined_main}), but also to quantify their alignment using correlation metrics (Table~\ref{tab:generation_correlations}). Through this, PROFILE reveals consistent patterns of agreement and misalignment that would be obscured by aggregate quality scores alone.
\textbf{Human and model preferences consistently misalign at the factor level across tasks}. While humans' most preferred factors vary by task, models consistently prioritize length across all tasks, suggesting models associate better quality with longer outputs. In both instruction-following tasks (Figure~\ref{fig:shp_main}) and document-based QA (Figure~\ref{fig:web_main}), humans prioritize $\mathsf{Receptiveness}$ and $\mathsf{Helpfulness}$. Although these two factors are also highly ranked for the models, the models always prioritize $\mathsf{Length}$ as the most important factor. 
% Humans in document-based QA also prioritize $\mathsf{Hallucination}$ avoidance, while models heavily emphasize $\mathsf{Length}$ and prioritize $\mathsf{Coherence}$ and $\mathsf{Helpfulness}$ more than humans do. 

\textbf{The misalignment pattern is particularly problematic in summarization tasks}. Humans prioritize $\mathsf{IntentAlignment}$, $\mathsf{FormalityAlignment}$, and $\mathsf{SourceCoverage}$ while penalizing the inclusion of words not in the original post, indicating the importance of maintaining the original content and style. In contrast, models consistently prefer longer summaries with new words (Table~\ref{tab:gpt_summaries_scores_5}). A full list of factor scores of all models across three tasks is available in the Appendix 
 (Table~\ref{Appendix:full_result_sum} - \ref{Appendix:full_result_webgpt}).
% \revise{Additionally, we analyze the correlations between features for each model~(Figure \ref{fig:corr_mat}) and find that each feature represents distinct aspects of model performance.}

To quantify this misalignment, we measure \textit{factor-level preference alignment} ($\tau$).
The left Generation column in Table~\ref{tab:generation_correlations}  shows that even the best-performing model (Gemini 1.5) only achieves a 0.289 $\tau$ correlation with human preferences in summarization. Similar low correlations are observed in the other two tasks (Table~\ref{tab:non_summary_evaluation_correlation}).  
This low correlation highlights the limitations of current models in capturing the granular aspects comprising human preference.

\textbf{Qualitative analysis demonstrates how PROFILE explains the observed misalignment.}
% Qualitative analysis further illustrates how our factor-level approach explains this misalignment. 
In a Reddit post below, GPT-4o's score 5 summary is longer and includes more facts than its score 3 summary, yet the shorter summary is human-preferred.
% (Table~\ref{tab:summary_gen_example_main_text}). 
The higher-scored model summary includes irrelevant details like ``Midwest hometown'' and ``new to Reddit,'' demonstrating the model's tendency to prioritize information quantity over relevance. Full examples are in Appendix~\ref{tab:summary_gen_example}.

\begin{tcolorbox}[colback=white,colframe=gray!50!black,title=GPT-4o Generation Sample, breakable]
\small
\textbf{Post:} Good Morning/Afternoon r/advice, Never posted on Reddit before at all, but I figured (based on the overall reliability of you nice individuals) (...) \\
\\
\textbf{Score 5} 
\textcolor{magenta}{[length: 93, \# facts: 10, src. coverage: 0.389]}: A Reddit user \textbf{recently moved back to their Midwest hometown} and, while setting up utilities for their new place, discovered they owe \$500 in gas bills from a college house they lived in until 2012. (...) \\
\\
\textbf{Score 3} \textcolor{magenta}{[length: 61, \# facts: 9, src. coverage: 0.44]}: A Reddit user seeks advice after discovering they owe \$500 in gas bills from a college house they left in 2012. (...) \textbf{\textcolor{blue}{(Human Preferred Output)}}
\end{tcolorbox}
\section{Achieving Better Alignment Using the Model as an Evaluator}
\label{sec:application}
In this section, we examine factor-level alignment in discrimination tasks. Specifically, we turn to the \textit{evaluation setting}, where models are used to determine which of two outputs is better.
Prior work has shown that model behavior can differ substantially between generation and discrimination tasks~\cite{west2023generative, oh-etal-2024-generative}. We therefore examine whether factor-level alignment also varies in evaluation (\S\ref{sec:discussion_eval}), and whether any observed improvements can be leveraged to guide better generation (\S\ref{sec:discussion_leveraging}).

% \paragraph{Experimental Setup.} 

\subsection{Factor-level Alignment in Evaluation}
\label{sec:discussion_eval}

% We use the same models as we use in \S~\ref{sec:uncovering_pref}. For each task, the model determines the preferred response between the two. We use the responses provided in the datasets. The specific prompts used for this evaluation are provided in the Appendix~\ref{sec:appendix_prompts}.
We examine the same models from \S\ref{sec:uncovering_pref} in an evaluation setting where models perform pairwise comparisons to determine which response is better. We use the response pairs provided in the existing datasets, with evaluation prompts detailed in Appendix\ref{sec:appendix_prompts}.

\begin{table}[htbp]
    \centering
    \small
    \begin{tabular}{@{}l|c|cc@{}}
        \toprule
        & \textbf{Gen.} & \multicolumn{2}{c}{\textbf{Eval.}} \\ \cmidrule(lr){2-2} \cmidrule(lr){3-4}
        & \textit{$\tau$} & \textit{$\tau$} & \textit{Agree. (\%)} \\ \midrule
        Mixtral       & 0.200 & 0.244 & 0.526 \\
        Tulu 2.5 + PPO (13B RM) & -0.156 & 0.511 & 0.516 \\
        Tulu 2.5 + PPO (70B RM) & 0.111 & 0.644 & 0.520 \\
        LLaMA 3.1 70B & 0.111 & 0.733 & 0.705 \\
        Gemini 1.5     & \textbf{0.289} & \underline{0.778} & \underline{0.721} \\
        GPT-4o         & 0.156 & \textbf{0.822} & \textbf{0.784} \\ 
        \bottomrule
    \end{tabular}
    \caption{Kendall's $\tau$ correlation in generation and evaluation settings, and evaluation agreement rate (\%) for the summarization task.}
    \label{tab:gen_eval_tau_agree}
\end{table}

Our analysis reveals that models demonstrate significantly stronger factor-level alignment with human preferences during evaluation compared to generation tasks. Table~\ref{tab:gen_eval_tau_agree} illustrates this pattern by comparing \textit{factor-level preference alignment} between humans and models, measured using Kendall's $\tau$. Alignment scores are consistently higher in evaluation across all models: GPT-4o achieves the highest evaluation alignment ($\tau = 0.82$) while showing substantially lower generation alignment ($\tau = 0.16$). This gap between evaluation and generation capabilities suggests that models possess stronger discriminative abilities than their generative performance would indicate.
% \input{Table/main_result_2_summ}
% \input{Table/Addition_SFT}
% These findings raise a natural question:
% % beyond agreement at the overall level, do LLM judges exhibit alignment with human preferences at the finer, factor level? And if so, 
% \textit{can we leverage this alignment to actively improve alignment in generation tasks?}

This substantial difference in alignment capabilities raises a natural question: \textit{can we leverage models' superior evaluation alignment to actively improve their generation alignment?}

\subsection{Leveraging LLM-as-an-evaluator}
\label{sec:discussion_leveraging}
Given the substantial gap between models' evaluation and generation alignment, we investigate whether LLMs' superior evaluation capabilities can be leveraged to improve generation performance. Using Reddit TL;DR summarization dataset, we explore two complementary approaches: self-refinement through supervised fine-tuning and feedback-driven generation.
\begin{table}[]
\small
\centering
\begin{tabular}{@{}l|lll@{}}
\toprule
                   & $\tau$   & $\rho$   & $r$ \\ \midrule
Tulu 2.5 w/o SFT & 0.111 & 0.2   & -0.015    \\
Tulu 2.5 human-SFT & -0.111 & -0.167 & -0.141 \\
Tulu 2.5 self-SFT          & \textbf{0.156} & \textbf{0.297} & \textbf{0.028}     \\ \bottomrule
% GPT-Eval SFT           & 0.156 & 0.285 & 0.007     \\ \bottomrule
\end{tabular}
\caption{Factor-level preference correlations between humans and Tulu 2.5 (70B RM) with and without supervised fine-tuning from self-evaluation (self-SFT).}
\label{tab:SFT_result}
\vspace{-10pt}
\end{table}
\subsubsection{Gen-Eval gap explains self-refinement's effectiveness}

\begin{table*}[ht]
\centering
\small
\begin{tabular}{@{}l|l*{6}{S[table-format=-1.2]}@{}}
\toprule
& \multicolumn{2}{c}{GPT-4o} & \multicolumn{2}{c}{LLaMA 3.1 70B} & \multicolumn{2}{c@{}}{Tulu 2.5 + PPO (70B RM)} \\
\cmidrule(lr){2-3} \cmidrule(lr){4-5} \cmidrule(l){6-7}
& {$\tau_G$} & {$\tau_H$} & {$\tau_G$} & {$\tau_H$} & {$\tau_G$} & {$\tau_H$} \\
\midrule
Baseline$_\text{A}$ & -0.24 & -0.07 & -0.20 & -0.29 & -0.29 & -0.29 \\
Baseline$_\text{B}$ & -0.29 & -0.29 & -0.42 & -0.42 & -0.24 & -0.24 \\
% \rowcolor{gray!20}
GPT-4o feedback & \textbf{ 0.36} & \textbf{\quad 0.45} & \textbf{\quad 0.29} & \textbf{\quad 0.20} & \textbf{\quad 0.16} & \textbf{\quad 0.16} \\
\bottomrule
\end{tabular}
\caption{Factor-level alignment (\(\tau\)) between improvements made by different generators (GPT-4o, LLaMA 3.1 70B, Tulu 2.5 + PPO (70B RM)) and factor-level preferences from GPT-4o (evaluation) and human. \(\tau_G\) and \(\tau_H\) indicate alignment with GPT-4o and human preferences respectively. Higher values show stronger alignment.}
%\revise{Kendall Tau Correlation Coefficients ($\tau$) between the delta Kendall scores of improved generations by different generators and evaluation factor preferences from GPT-4o and human. $\tau_G$ represents the correlation with GPT-4o preferences, while $\tau_H$ represents the correlation with human preferences. Higher coefficients indicate that changes in factor-level alignment are more closely aligned with the fine-grained preferences of GPT-4o or human evaluators.}
\label{tab:bestofn}

\end{table*}
 
We investigate whether supervised fine-tuning (SFT) with self-evaluation can bridge the alignment gap in generation tasks. Using TULU 2.5 (70B RM), we generate summaries with target scores 1-5, then use the same model to perform pairwise evaluations and re-rank these summaries based on win rates. 
We then fine-tune the generator on 4,000 such examples, where inputs are instructions to generate summaries of different quality levels (1-5) and outputs are the re-ranked summaries. We evaluate the fine-tuned model on 500 unseen examples.
% The generator is then SFT-trained on 4,000 such examples and tested on 500 unseen examples. The input is an instruction to generate summaries of scores 1-5 given a post, and output labels are the re-ranked summaries of score 1-5. 

Table~\ref{tab:SFT_result} demonstrates that this self-evaluation approach significantly improves generation alignment, achieving performance comparable to GPT-4o (see Table~\ref{tab:generation_correlations}). Notably, a baseline model trained on human-preferred responses (\textit{human-SFT}) using identical training configurations actually performed worse than the original TULU model. This counterintuitive result aligns with broader observations that even DPO- or RLHF-trained models often struggle to consistently align with human preferences. This result highlights the potential of leveraging models' evaluation capabilities for training.

% Table~\ref{tab:SFT_result} shows that the SFT-trained model achieves significantly improved alignment compared to the original TULU model, reaching performance comparable to GPT-4o (Table~\ref{tab:generation_correlations}). 
% As another baseline, we consider a model trained on human-preferred responses (\textit{human-SFT}) using the same post samples and training configuration. This baseline closely resembles widely used DPO-style preference optimization. Interestingly, the human-SFT model performed worse than the original TULU without SFT, showing lower correlation with human preferences. This result aligns with broader observations that even DPO- or RLHF-trained models often struggle to consistently align with human preferences, underscoring the effectiveness of the PROFILE framework in providing guidance for training signals.

% \input{Table/Addition_bon} 
\subsubsection{Leveraging evaluation for better alignment in generation}
% Prior analysis indicates stronger factor-level alignment during evaluation than generation (Section~\ref{sec:eval_alignment}), suggesting that evaluator feedback might improve generation alignment.

We further explore whether explicit evaluator feedback can improve generation alignment in real-time. Our approach involves a generator producing two initial summaries per post, followed by an evaluator selecting the preferred response (or tie) and providing a detailed justification. The generator then uses this feedback to produce an improved summary.

Using GPT-4o as the evaluator, we compare this feedback-driven approach against two baselines: (1) \textit{Baseline$_A$}: the generator produces one improved summary from both initial summaries without external feedback; (2) \textit{Baseline$_B$}: the generator produces two improved summaries without feedback, each refined from one initial summary. These baselines represent typical self-improvement scenarios. We evaluate on 100 samples across three generators: GPT-4o, LLaMA 3.1 70B, and Tulu 2.5 + PPO.

Table~\ref{tab:bestofn} shows that incorporating evaluator feedback consistently improves alignment with both GPT-4o and human judgments across all generators. In contrast, both baselines show negative correlations, indicating that implicit self-critique without explicit feedback (\emph{i.e.,} re-generation) actually diverges from desired preferences. Manual analysis of 30 samples confirms that evaluator feedback effectively emphasizes factors that align with evaluation preferences (see Appendix~\ref{sec:appendix-best-of-n-feedback-valid} for detailed analysis).

These findings demonstrate that leveraging models' superior evaluation capabilities---either through self-refinement during training or explicit feedback during generation---can effectively improve factor-level alignment in generation tasks. See Appendix~\ref{sec:appendix-best-of-n-prompts} for prompt and metric details.
\section{Related Work}
% \paragraph{Meta Evaluation of LLMs}
\paragraph{Human-AI Preference Alignment.}
Aligning LLMs with human preferences is a central focus in LLM research, leading to techniques like supervised instruction tuning~\citep{mishra2021cross, wei2021finetuned}, RLHF~\citep{ouyang2022training}, DPO~\citep{guo2024direct}, and RLAIF, which utilizes AI-generated feedback~\citep{bai2022constitutional, Lee2023RLAIFVR}. However, most studies focus on overall performance (e.g., a response as a whole).
% but most focus on overall performance or high-level objectives. 
% or instruction tuning, various techniques have shown effectiveness such as task and model scaling (Mishra et al., 2022; Wei et al., 2021; Wang et al., 2022c; Chung et al., 2022),
While some work has explored using fine-grained human feedback~\citep{dong2023steerlm, wu2024fine}, a comprehensive understanding of how granular factors contribute to and differentiate human and model preferences is still lacking. \citet{hu-etal-2023-decipherpref} address this gap by deciphering the factors influencing human preferences. 
We extend this work by analyzing factor-level preferences across multiple tasks and comparing the driving factors of both humans and model preferences.
% Building on this work, we expand the investigation of factor-level preference across multiple tasks and extend the preference analysis to models, providing a comparative analysis of the factors driving both human and model preferences.
% \revise{Similar to fine-grained RLHF \citep{wu2023finegrainedhumanfeedbackgives}, we can leverage factor-level scores to align the LLM. Additionally, similar to LLMRefine \citep{xu-etal-2024-llmrefine}, we can employ fine-grained guidance to harness the LLM's self-refinement capability for further improvement.}

\paragraph{Fine-grained Evaluation of LLMs.}
Recent research has increasingly emphasized the need for more fine-grained evaluations of LLMs. For instance, researchers have proposed fine-grained atomic evaluation settings for tasks like fact verification and summarization~\citep{min2023factscore, krishna-etal-2023-longeval}, developed a benchmark for fine-grained holistic evaluation of LLMs on long-form text~\citep{ye2024flask}, and enhanced evaluation transparency through natural language feedback~\citep{xu-etal-2023-instructscore}. Building on this trend, our work shifts from evaluating individual factors in isolation to analyzing their influence on human preferences and investigating the alignment between human and model judgments regarding the relative importance of these factors.

\paragraph{Analyzing behaviors of LLM-as-a-judge.}
Furthermore, researchers are actively exploring the potential of LLMs as evaluators. \citet{fu2024gptscore, madaan2024self, liu2023g} demonstrate the capacity of large models like GPT-4 to achieve human-level evaluation. 
% \citep{xu-etal-2023-instructscore} explores LLM evaluator under fine-grained error annotations. 
However, recent works reveal discrepancies in model performance between generation and evaluation tasks~\citep{west2023generative, oh-etal-2024-generative}. Inspired by frameworks to meta-evaluate LLM as an evaluator~\citep{zheng2023judging, ribeiro-etal-2020-beyond}, our work evaluates not only the quality of model-generated text but also the alignment of model preferences in evaluation settings, providing a more comprehensive assessment of LLM capabilities.
\section{Conclusion}
We introduce PROFILE, a framework for granular factor level analysis of LLM alignment with human preferences. Our analysis using PROFILE reveals that LLMs tend to over-prioritize factors like output length, misaligning human preferences during generation. However, these models exhibit stronger alignment in evaluation tasks, indicating the potential for leveraging evaluative insights to improve generative alignment. PROFILE facilitates a nuanced understanding of the alignment gaps between human and model preferences. These insights underscore the necessity for more sophisticated, factor-level alignment strategies that can guide the development of LLMs to better align with human expectations, ultimately fostering more reliable aligned AI systems.
% \newpage
\section{Limitations}
% \noindent\textbf{Limitations.}
This study has several limitations. First, the preference datasets used may not fully represent the entire spectrum of human preferences. Second, due to budget constraints, human evaluations of model outputs were conducted on a limited scale, with a restricted number of participants, and only on one task. Furthermore, this study represents a preliminary exploration into methods for achieving better alignment, highlighting the potential of various techniques to enhance generation and evaluation. Extensive studies are required to thoroughly assess the efficacy and generalizability of these methods. While this study focuses on post-hoc correction methods, future research should investigate how to incorporate the identified preference factors as signals during the training stage. Additionally, exploring how to embed these signals within datasets used for preference optimization represents a promising direction for future work.
\section{Ethics Statement}
Our research relies on established benchmarks and models, and does not involve the development of new data, methodologies, or models that pose significant risks of harm. The scope of our experiments is limited to analyzing existing resources, with a focus on model performance. Human studies conducted within this work adhere to relevant IRB exemptions, and we ensure fair treatment of all participants. Our work is mainly focused on performance evaluation, we recognize that it does not specifically address concerns such as bias or harmful content.

% \section{Reproducibility Statement}
% \input{Text/9_Reproducibility}

\section*{Acknowledgments}
This work was supported by Institute for Information \& communications Technology Promotion(IITP) grant funded by the Korea government (MSIP) (No. RS-2024-00443251, Accurate and Safe Multimodal, Multilingual Personalized AI Tutors)

\bibliography{custom}
\clearpage
\appendix

\section*{Appendix}
\label{sec:appendix}
\section{Experimental Setting}
\subsection{Tasks}
\label{appendix:sec_task}
We examine three publicly available datasets of pairwise human judgments commonly used in preference optimization methods like RLHF and DPO training: 
\textbf{Reddit TL;DR} We analyze the dataset released by OpenAI~\citep{stiennon2020learning}, which includes human ratings of summaries across multiple axes (referred to as ``\texttt{axis evaluations}''). Higher scores indicate human preference across multiple evaluation dimensions.
\textbf{StanfordHumanPreference-2 (SHP-2)}~\citep{pmlr-v162-ethayarajh22a}, focuses on capturing human preferences over responses to questions and instructions, prioritizing helpfulness. Higher scores indicate a more helpful response. For this study, we use responses from the ``\texttt{reddit/askacademia}'' domain.
\textbf{OpenAI WebGPT} This dataset~\citep{nakano2021webgpt}, addresses the task of generating answers to questions from the ELI5 (\textit{``Explain Like I’m Five''}) subreddit. Human annotations compare two model-generated answers based on factual accuracy and overall usefulness.
% We examine three publicly available datasets of pairwise human judgments commonly used in preference optimization methods, such as RLHF and DPO training. For summarization task, we analyze the dataset released by OpenAI~\citep{stiennon2020learning}, which includes human ratings of summaries along multiple axes, referred to as ``\texttt{axis evaluations}'' for the \textbf{Reddit TL;DR dataset}. In this dataset, higher scores indicate the preferred summary across various evaluation dimensions. 
% The \textbf{StanfordHumanPreference-2 (SHP-2)}~\citep{pmlr-v162-ethayarajh22a} focuses on capturing human preferences over responses to questions and instructions, with an emphasis on evaluating helpfulness. Human judgments reflect which response is more helpful, with higher scores signifying the preferred response. Within this dataset, we specifically use responses from the ``\texttt{reddit/askacademia}'' domain. Lastly, \textbf{OpenAI WebGPT}~\citep{nakano2021webgpt} is used for factual and open-ended question-answering. This dataset addresses the task of generating answers to questions from the ELI5 ``Explain Like I’m Five'' subreddit. Human annotations in this dataset compare two model-generated answers per question, evaluating them based on factual accuracy and overall usefulness. 
We exclude pairs with Tie ratings in all three datasets, as our analysis focuses on cases with clear preference distinctions.

\subsection{Models}
\label{appendix:sec_model}
Our study focuses on the most advanced and widely-used generative models currently accessible, encompassing both proprietary and open-source options. For open-source models, we include LLaMA 3.1 70B~\citep{dubey2024llama}\footnote{Inference for LLaMA was conducted using the Together AI API. ~\url{https://www.together.ai/}}, Mixtral 8x7B Instruct v0.1~\citep{jiang2024mixtral}, three TÜLU 2.5 Models~\citep{ivison2024unpacking}---TÜLU 2.5 + PPO 13B (13B RM)~\footnote{We use huggingface allenai/tulu-v2.5-ppo-13b-uf-mean-13b-uf-rm model.}, TÜLU 2.5 + PPO 13B (70B RM)~\footnote{We use huggingface allenai/tulu-v2.5-ppo-13b-uf-mean-70b-uf-rm model.}, and TÜLU 2.5 + DPO 13B~\footnote{We use huggingface allenai/tulu-v2.5-dpo-13b-uf-mean model.}. For proprietary models, we use Gemini 1.5 Flash~\citep{reid2024gemini}, GPT-4o~\citep{openai2024hello_gpt4o}~\footnote{We use gpt-4o-2024-05-13 version for all GPT-4o inference.}, and GPT-3.5~\footnote{We use gpt-3.5-turbo-1106 version for all GPT-3.5 inference.}.
We set the parameters for all models to: temperature = 0.6, top\_p = 0.9, and max\_tokens = 1024.
4 Quadro RTX 8000 48GB were used with CUDA version 12.4 when running TULU Models.

We used autrotrain library\thinspace\footnote{\url{https://huggingface.co/autotrain}} for supervised fine-tuning TULU model in experiments in \S~\ref{sec:application}. The parameters for fine-tuning are as follows:\\
  block\_size: 2048,
  model\_max\_length: 4096,
  epochs: 2,
  batch\_size: 1,
  lr: 1e-5,
  peft: true,
  quantization: int4,
  target\_modules: all-linear,
  padding: right,
  optimizer: paged\_adamw\_8bit,
  scheduler: linear,
  gradient\_accumulation: 8,
  mixed\_precision: bf16,
  merge\_adapter: true

\label{sec:appendix_task_model}

\begin{figure}[t]
    \centering
    \includegraphics[width=\columnwidth]{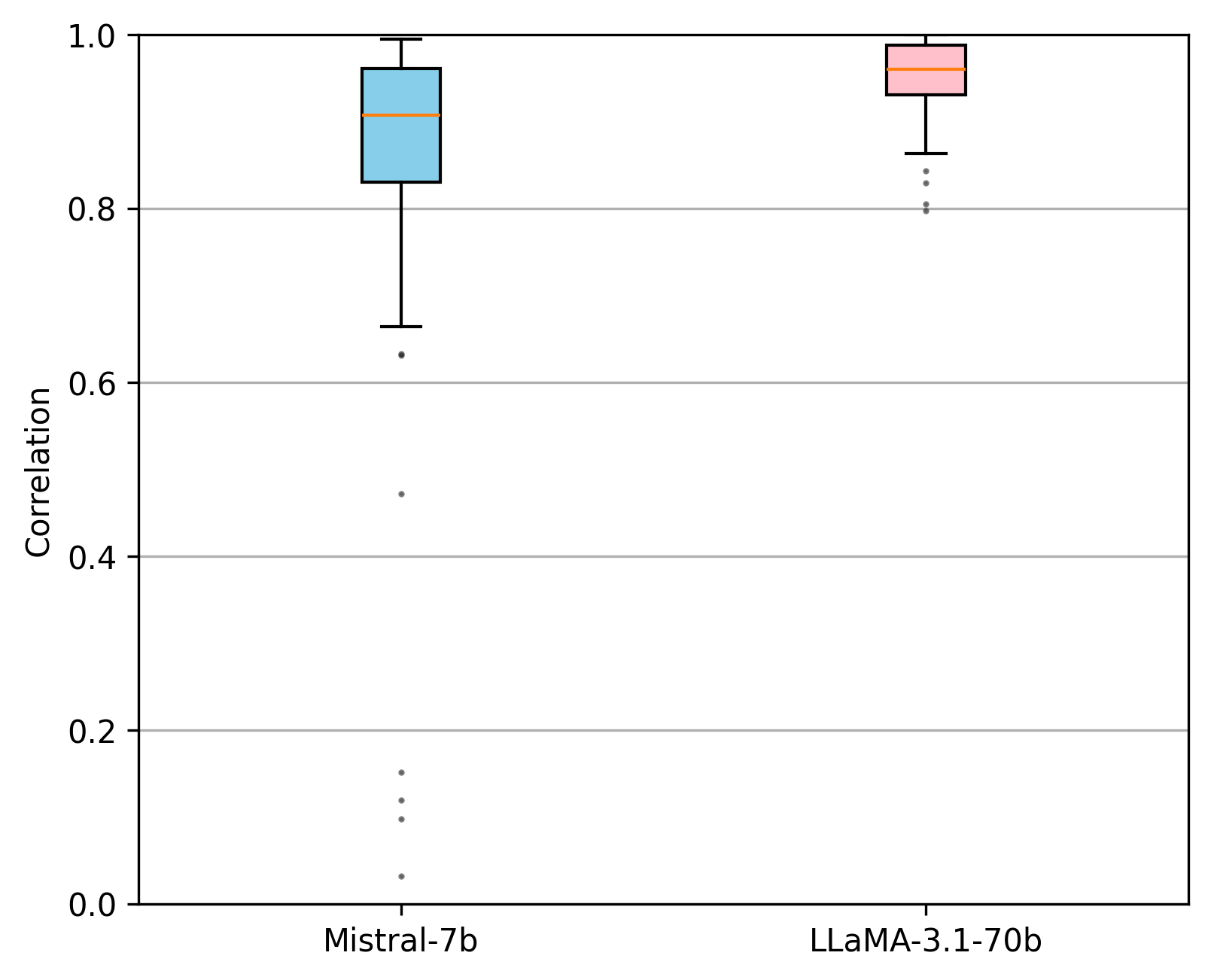}
    \caption{Pearson correlation between target conditioning scores and log probabilities of generated summaries for Mistral-7b and LLaMA-3.1-70b.}
    \label{fig:method-valid}
    \vspace{-3mm}
\end{figure}
\clearpage
\begin{figure*}[!tbhp]
    \centering
    \includegraphics[width=\textwidth]{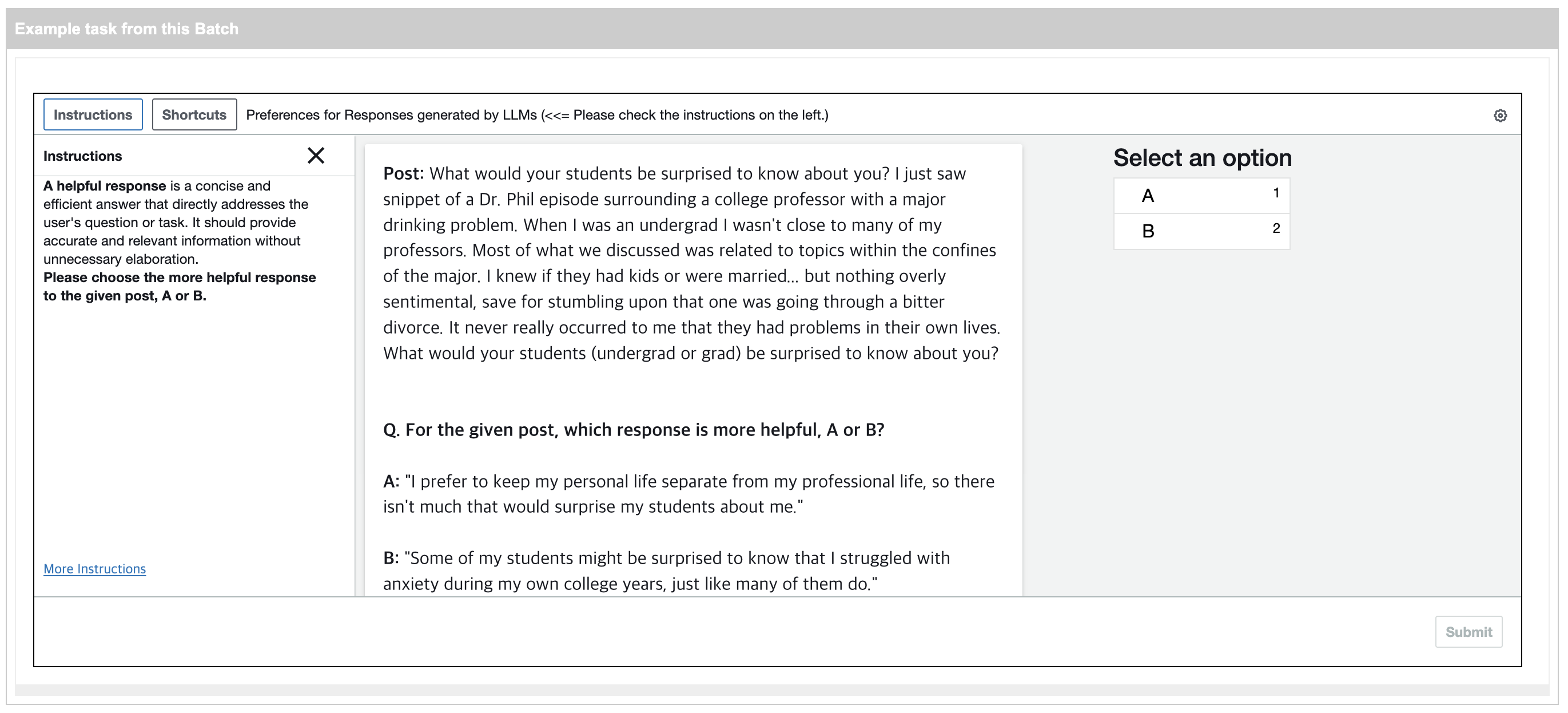}

    \caption{ A screenshot of a sample summary with preference annotations.}
    \label{appendix:mturk_screenshot}
\end{figure*}
\section{Human Evaluation of Model Generations}
We collect human preference data via Amazon Mechanical Turk (MTurk) for 30 posts and 6 models. For each post, three summary pairs—selected from five model-generated summaries (scored 1 to 5)—are presented to three annotators. Annotators, restricted to US-based workers with a 95\% and HIT approval rate and over 5,000 approved HITs are recruited. 
The MTurk task description clearly explained the study’s purpose and data usage. As shown in Figure~\ref{appendix:mturk_screenshot}, we provide detailed instructions about the experiment through MTurk, and participants who consented then participated in the study.

For the main experiment, we gave annotators the following instructions: \textit{``A good summary is a shorter piece of text that captures the essence of the original. It aims to accomplish the same purpose and convey the same key information as the original post. Please choose the better summary, A or B.''} Each annotation is compensated at \$0.25. This process yields 1,620 annotations (30 posts * 6 models * 3 pairs/model/post * 3 annotations/pair).

For each model and corresponding summary pair, we calculate an agreement rate, which indicates the percentage of pairs where at least two out of three annotators prefer the summary with the higher score assigned by the model. Each model is evaluated on 90 summary pairs, derived from 30 posts with three pairs per post. 
\label{sec:human_annotation}

% 이 table에서 factor-level preference alignment ($\tau$)는 매우 낮음에도, overall agreement는 56에서 75 퍼센트 range 로 꽤 높은 것을 확인할 수 있다. 모델이 만들어놓은 답을 보고 결과론적으로 이렇게 binary로 뭐가 더 좋은지 평가하게 하는 세팅으로는 이런 factor-level misalignment를 알아채기 어렵게 한다는 것을 알 수 있다. overall하게는 70점대 가까이 대부분 나오지만 사실 그 안에서 length나 다른 요소에 대한 불호(?)를 disentangle 하지 못하게 된다.
% 그러니까 지금 arena 방식으로 채택된 모델의 좋은 답은 언제나 더 긴 답이 될거고, 이게 human agreement 높은 학습 데이터로 사용되면 현재 모델에서 나타나는 verbose answer에 대한 선호가 계속 유지될 것이다. 

\subsection{Examples of Human-Model Misalignment}
\label{sec:appendix_example}
\begin{tcolorbox}
[colback=white,colframe=gray!50!black,title=GPT-4o misaligned examples, breakable]
\small
\textbf{Post:} Good Morning/Afternoon r/advice, Never posted on Reddit before at all, but I figured (based on the overall reliability of you nice individuals) that now would be a good time to start. \\\\
 About a year ago I moved away from my Midwest town to the west coast for work, and have just recently moved back. I was calling the Utility providers in my area to set up the gas/water/elec for my new place, but I was informed that gas could not be set up due to owing a past amount. \\\\
 % I am generally good about keeping up on those things, but am only human so I assumed I would have a small sum to pay. Maybe one months worth that I missed at the end of a lease or something like that. I was alarmed however when I was told that the value was around \$500.00 from a home that I shared with friends in college. They stated that this amount was accrued in 2013. \\\\
(...) \\\\
 My question is, am I really on the hook for these fees and what recourse could I possibly have? I cannot afford to pay it, but I also don't think that it should be on me TO pay it. Any advice is truly appreciated as I have not had any experience with this type of thing before. \\\\
\textbf{GPT-4o's score 5 generation} \textcolor{magenta}{[length: 93, \# facts: 10, src. coverage: 0.389]}: A Reddit user recently moved back to their Midwest hometown and, while setting up utilities for their new place, discovered they owe \$500 in gas bills from a college house they lived in until 2012. The user didn't cancel the service when they moved out, leading to the bill accruing over time. They are looking for advice on whether they are responsible for this debt and what recourse they might have, as they are unable to afford the payment. The user is new to Reddit and appreciates any guidance on handling this situation. \\\\
\textbf{GPT-4o's score 3 generation} \textcolor{magenta}{[length: 61, \# facts: 9, src. coverage: 0.44]}: A Reddit user seeks advice after discovering they owe \$500 in gas bills from a college house they left in 2012. They didn't cancel the service when they moved out, and the bill continued to accrue. They want to know if they are responsible for paying it and what actions they can take, as they can't afford to pay the bill. \textbf{\textcolor{blue}{(Human Preferred Output)}}
\label{tab:summary_gen_example}
\end{tcolorbox}
A higher score summary generated by GPT-4o, as shown by the factor-level analysis (in magenta), is longer and includes details not necessary to understand the essence of the original post.

\subsection{Human Evaluation Results of Model Generations}
The agreement rates for each model are summarized in Table~\ref{tab:gen_accuracy}.
Despite low factor-level preference alignment ($\tau$), overall agreement rates range from 56\% to 75\%. This suggests that binary evaluations, where annotators choose the better summary based on overall quality, can mask factor-level misalignments such as preferences for summary length. Since models consistently favor longer responses, continued reliance on this setup to generate human-annotated training data may reinforce the tendency towards verbose answers, aligning with human preferences superficially but missing finer-grained alignment on factors like length or conciseness.
% \begin{table}[]
% \centering
% \small
% \begin{tabular}{lcc}
% \hline
%                        & \textit{$\tau$} & \textit{Agree.(\%)}  \\ \hline
% Gemini 1.5 Flash       & \textbf{0.289}      & 0.689          \\
% Mixtral 8x7B Instruct  & {\ul 0.2}           & 0.678 
% \\
% GPT-3.5                & {\ul 0.2}           & {\ul 0.633}     \\
% GPT-4o                 & 0.156               & 0.722          \\
% Llama 3.1 70B          & 0.111               & 0.756          \\
% Tulu 2.5 PPO 13B (70B) & 0.111               & 0.567              \\
% % tulu-2.5-dpo-13b       & 0.111               & -              \\
% % tulu-2.5-ppo-13b-13brm & -0.156              & -          \\ 
% \hline
% \end{tabular}
% \caption{Fine-grained kendall correlation and the overall agreement with human preferences in Summary Generation Task. For Tulu 2.5, the size in the parentheses is the size of the RM used
% to train the LLM.}
% \label{tab:gen_accuracy}
% \end{table}

\begin{table}[thbp]
% \begin{wraptable}{r}{0.5\textwidth} 
% \vspace{-4mm}
\centering
\small
\begin{tabular}{lcc}
\hline
                       & \textit{$\tau$ } & \textit{Agree.(\%)}  \\ \hline
Gemini 1.5       & \textbf{0.289}      & 0.689          \\
Mixtral  & {\ul 0.2}           & 0.678          \\
% GPT-3.5                & {\ul 0.2}           & {\ul 0.633}    \\
GPT-4o                 & 0.156               & {\ul 0.722}          \\
LLaMA 3.1 70B          & 0.111               & \textbf{0.756}          \\
Tulu 2.5 + PPO (70B RM) & 0.111               & 0.567          \\
\hline
\end{tabular}
\caption{Factor-level preference alignment ($\tau$) and response-level agreement with human preferences in the summarization Task.}
\label{tab:gen_accuracy}
% \vspace{-4mm}
% \end{wraptable}
\end{table}
% \clearpage
% \clearpage
{\section{Additional Examples of Model Generations and Evaluations}}
{\subsection{Generation}}

\begin{table*}[!htpb]
\small

\begin{tabular}{@{}>{}p{0.17\textwidth} >{}p{0.17\textwidth} >{}p{0.17\textwidth} >{}p{0.19\textwidth} >{}p{0.2\textwidth}@{}}

\toprule
\multicolumn{1}{c}{\textbf{Score 1}} & 
\multicolumn{1}{c}{\textbf{Score 2}} & 
\multicolumn{1}{c}{\textbf{Score 3}} & 
\multicolumn{1}{c}{\textbf{Score 4}} & 
\multicolumn{1}{c}{\textbf{Score 5}} \\

\midrule
The poster and their fiance have scabies and are treating it. Their roommate refuses to get treated or clean his things, risking reinfestation. The poster is considering asking him to move out. &
A couple living with a roommate got scabies and started treatment. The roommate, who has no symptoms, refuses to treat himself or his belongings. The poster is worried about reinfestation and may ask the roommate to move out since he isn't on the lease. &
A 19-year-old woman and her 20-year-old fiance contracted scabies after moving into a new apartment with a mutual friend. Both are undergoing treatment, but the roommate refuses to get treated or clean his belongings. The poster is concerned about reinfestation and is contemplating asking the roommate to move out, as he is not on the lease. &
A 19-year-old woman and her 20-year-old fiance recently moved into a new apartment with a mutual friend and contracted scabies. Despite both of them undergoing treatment, their roommate refuses to take any preventive measures or treat himself, even though he has not shown symptoms yet. The poster is worried that this will lead to a reinfestation and is considering asking the roommate to move out since he is not on the lease. &
A young woman (19) and her fiance (20) recently moved into a new apartment with a mutual friend and unfortunately contracted scabies. Both have started treatment after a doctor's confirmation. However, their 20-year-old roommate, who has not shown symptoms, refuses to undergo treatment or clean his clothes and linen. This refusal has caused concern for the woman, as she fears reinfestation and is seriously considering asking the roommate to move out, especially since he is not on the lease. \\
\bottomrule
\end{tabular}

\caption{{An example of GPT-4-generated summaries of scores 1-5.}}
\label{tab:gpt_summaries_scores_5}
\end{table*}
{Table~\ref{tab:gpt_summaries_scores_5} shows GPT-4o generated summaries of scores 1-5. There is a clear trend of higher score summaries being longer. The information additional to lower-scoring summaries is the ages of the characters, which is not essential in understanding the original post. This is a randomly selected example and a typical model-generated summary case.}

{Additionally, we analyze cases where model-generated outputs receive higher scores but are not preferred by human annotators (based on majority votes). The human-model misalignment is more pronounced when comparing summaries with larger score gaps assigned by the models. 
Below, we provide additional examples of such divergence for GPT-4o, LLaMA 3.1 70B and Tulu 2.5 + PPO (13B RM).
We find that when the length and information gap is significant, humans favor summaries that are shorter and less information-dense, as model-generated summaries with lower scores (1-2) generally convey the main intention of the post as well. When factual content is similar, humans tend to favor more informal responses (e.g., ``Guy'' over ``An 18-year-old male'', and ``gotten back together'' over ``renewed romantic connection''), which aligns with our factor-level findings indicating higher formality-alignment for human-preferred outputs, since Reddit posts are mostly informal.}

{\textit{\textcolor{red}{Warning: Examples below contain content that may be offensive or upsetting.}}}

\begin{tcolorbox}
[colback=white,colframe=gray!50!black,title=GPT-4o misaligned examples,breakable]
\small
\textbf{GPT-4o's score 4 generation}: An 18-year-old male in a long-distance relationship with his 17-year-old girlfriend told her he loves her but wants an open relationship. The girlfriend, who has been with four other boys before him, reacted very negatively, telling him to ``FUCK OFF" and that she never wants to speak to him again. He is seeking advice on how to make the situation easier for her and whether he can maintain their relationship while exploring other people. \\\\
\textbf{GPT-4o's score 1 generation}: Guy told his girlfriend he wants an open relationship. She got mad and told him to go away. He wants advice. \textbf{\textcolor{blue}{(Human Preferred Output)}}
\end{tcolorbox}

\begin{tcolorbox}
[colback=white,colframe=gray!50!black,title=LLaMA-3.1-70b misaligned examples]
\small
\textbf{LLaMA's score 4 generation}: I'm in a relationship with an 18-year-old girl who's a self-described Tsundere. We share similar interests, but her mood swings are causing problems. She can be affectionate and loving, but also violent and aggressive, lashing out at me over minor issues. For example, she threw shoes at me for a joke and almost stabbed me with a knife over a trivial matter. I'm scared and unsure how to navigate these outbursts. \\\\
\textbf{LLaMA's score 2 generation}: I'm dating a girl who's similar to me, but she's really violent and has mood swings. She gets angry over small things and has even attacked me. I'm scared and need help.
\textbf{\textcolor{blue}{(Human Preferred Output)}}
\end{tcolorbox}

\begin{tcolorbox}
[colback=white,colframe=gray!50!black,title=Tulu 2.5 + PPO (13B RM) misaligned examples]
\small
\textbf{Tulu's score 4 generation}: The OP discovered a potentially incriminating text exchange between their SO and her ex's sister, suggesting a renewed romantic connection. \\\\
\textbf{Tulu's score 1 generation}: The OP read a suspicious text on their SO's phone, sent from her ex's sister, that suggested they might have gotten back together.
\textbf{\textcolor{blue}{(Human Preferred Output)}}
\end{tcolorbox}
{\subsection{Evaluation}}
{We provide examples where the model evaluations align with human preferences, even if the chosen option contains less facts or is shorter. In the first example, where both GPT-4o and LLaMA 3.1 70B correctly chose human-preferred summary, while the chosen summary is shorter, it more accurately reflects the key issue in the original post by mentioning the writer's economic status. In the second example, the GPT-4o chosen summary is more clearly reflecting the content in post over the other option which analogically describes the main idea of the post.}

\begin{tcolorbox}
[colback=white,colframe=gray!50!black,title=GPT-4o \& LLaMA aligned examples, breakable]
\small
\textbf{Post:} Yesterday, I accidentally dropped my Motorola Atrix 2 and the screen cracked really badly. My phone is still fully functional, but it's a bit difficult to see what I'm doing when I'm texting or web browsing, etc. Anyway, I stupidly didn't buy insurance for my phone and I'm not eligible for an upgrade until next May! AT\&T offers some options as far as getting a no-commitment phone at a slight discount, but spending \$300-\$600 for a new phone isn't really in the budget right now. \\
(...) \\
I found a couple websites that will repair your phone if you send it in. [Doctor Quick Fix] will do it for \$110 and I'm still waiting on a quote from [CPR](So my question is, have any of you used this company, or know anyone who has used it? Should I trust these companies? Do you have any recommendations? What should I do to get my phone fixed? \\\\
\textbf{Summary A}: Dropped my phone, they said they won't repair phones that have been physically abused. Looking for suggestions on cell phone repair companies, if any, and what I should do to get my phone fixed. \\\\
\textbf{Summary B}: I dropped my phone, cracking the screen. I can't afford to buy a full price phone, so should I try the above repair companies? What should I do? \textbf{\textcolor{blue}{(Human Preferred Output)}}
\end{tcolorbox}

\begin{tcolorbox}
[colback=white,colframe=gray!50!black,title=GPT-4o aligned \& LLaMA misaligned examples,breakable]
\small
\textbf{Post:} I got a letter in the mail saying I've been passed up for being hired for my dream job. I wanted this job for 10 damn years and now it's over. I've trained my body, mind, and soul for this job and just through a simple letter, I've been removed from that process. I was in good standing with getting hired. Passed everything with flying colors. \\
(...) \\
Now what? Am I to live with my parents the rest of my life? Am I to never get my dream car? Am I to just keep my job where I only get paid minimum wage while I make the company tens of thousands? I don't know what to do. I mean my second dream job would be to work with penguins, but I don't think that's possible for me. Anyone have any advice for me? What should I do? \\\\
\textbf{Summary A}: I followed the yellow brick road  for half my life and ended up at a complete dead end and I can't turn around to go back. \\\\
\textbf{Summary B}: Got passed up for a dream job. Now what the hell are I supposed to do with my life that doesn't include my dream job? \textbf{\textcolor{blue}{(Human Preferred Output)}}
\end{tcolorbox}

\begin{tcolorbox}
[colback=white,colframe=gray!50!black,title=GPT-4o misaligned examples]
\small
\textbf{GPT-4o's score 4 generation}: An 18-year-old male in a long-distance relationship with his 17-year-old girlfriend told her he loves her but wants an open relationship. The girlfriend, who has been with four other boys before him, reacted very negatively, telling him to ``FUCK OFF" and that she never wants to speak to him again. He is seeking advice on how to make the situation easier for her and whether he can maintain their relationship while exploring other people. \\\\
\textbf{GPT-4o's score 1 generation}: Guy told his girlfriend he wants an open relationship. She got mad and told him to go away. He wants advice. \textbf{\textcolor{blue}{(Human Preferred Output)}}
\end{tcolorbox}

\begin{tcolorbox}
[colback=white,colframe=gray!50!black,title=Llama-3.1-70b misaligned examples]
\small
\textbf{Llama's score 4 generation}: I'm in a relationship with an 18-year-old girl who's a self-described Tsundere. We share similar interests, but her mood swings are causing problems. She can be affectionate and loving, but also violent and aggressive, lashing out at me over minor issues. For example, she threw shoes at me for a joke and almost stabbed me with a knife over a trivial matter. I'm scared and unsure how to navigate these outbursts. \\\\
\textbf{Llama's score 2 generation}: I'm dating a girl who's similar to me, but she's really violent and has mood swings. She gets angry over small things and has even attacked me. I'm scared and need help.
\textbf{\textcolor{blue}{(Human Preferred Output)}}
\end{tcolorbox}

\begin{tcolorbox}
[colback=white,colframe=gray!50!black,title=Tulu 2.5 + PPO (13B RM) misaligned examples]
\small
\textbf{Tulu's score 4 generation}: The OP discovered a potentially incriminating text exchange between their SO and her ex's sister, suggesting a renewed romantic connection. \\\\
\textbf{Tulu's score 1 generation}: The OP read a suspicious text on their SO's phone, sent from her ex's sister, that suggested they might have gotten back together.
\textbf{\textcolor{blue}{(Human Preferred Output)}}
\end{tcolorbox}

% \input{Table/Appendix_gpt_summary_full_scores}
% \clearpage

\section{PROFILE}
\label{sec:appendix_profile}
\subsection{Validation }
Figure~\ref{fig:method-valid} shows the distribution of Pearson correlations over 100 samples for both LLaMA-3.1-70B and Mixtral.

We find that the correlation of most samples are concentrated between 0.85 and 1.0, indicating a strong correlation between the target scores in our score-conditioned setting and the models' log probabilities (i.e., their preference for those responses)

\subsection{Factor Extraction Methods}
\paragraph{Rule-based Extraction}
We obtain the $\mathsf{Length}$ and $\mathsf{Novel\:Words}$ using a rule-based extraction method. First, we calculate the output's length and count the novel words by removing special characters and splitting the text into words. The total word count represents $\mathsf{Length}$. For $\mathsf{Novel\:Words}$, we stem both the source text and the model output to create unique sets of stemmed words, then determine the number and proportion of unique words in the output that differ from the source.

\paragraph{LLM-based Extraction}
The calculations are divided into atomic-fact-level and response-level based on the granularity of the factors.

Atomic-Fact-Level Factors refer to those factors that are evaluated based on the presence or absence of each factor at the atomic fact level. An atomic fact is a short, self-contained piece of information that does not require further explanation and cannot be broken down further~\citep{min2023factscore}. These include the  $\mathsf{Number\:Of\:Facts}$, $\mathsf{Source\:Coverage}$, $\mathsf{Off\:Focus}$, $\mathsf{Hallucination}$, $\mathsf{Helpfulness}$, and $\mathsf{Misinformation}$. The $\mathsf{Number\:Of\:Facts}$ is determined by counting the total atomic facts, while the remaining factors are calculated as the ratio of relevant atomic facts to the total number of atomic facts.

Response-Level Factors refer to those factors that are evaluated based on the presence or absence of each factor at the response level. These include $\mathsf{Receptiveness}$, $\mathsf{Intent\:Alignment}$, and $\mathsf{Formality\:Alignment}$. $\mathsf{Formality\:Alignment}$ is classified into one of three categories: [Aligned/Misaligned/Partially-Aligned], while the other two factors are determined in a binary manner [Yes/No].

The prompts used are provided in \ref{sec:Appendix-factor-prompt}. The $\mathsf{Source\:Coverage}$ does not have a separate prompt since it was calculated using the output from the $\mathsf{Hallucination}$  (i.e., the ratio of non-hallucinated atomic facts to the total number of atomic facts in the Source Post).

{\paragraph{Cost of LLM-based Extraction.} Here we report the average cost required for LLM-based extraction using GPT-4o. Table~\ref{tab:llm_cost} shows the average cost for each factor in a single sample of the summarization task, with the total cost being \$0.018 per post sample.}

\begin{table}[t]
\centering
\resizebox{\columnwidth}{!}{%
\begin{tabular}{lccc}
\toprule
\textbf{Factor} & \textbf{Input} & \textbf{Output} & \textbf{Sum} \\
\midrule
Atomic facts        & \$0.00146 & \$0.00057 & \$0.00203 \\
Hallucination       & \$0.00165 & \$0.00203 & \$0.00368 \\
Off-focus           & \$0.00332 & \$0.00236 & \$0.00568 \\
Intent-alignment    & \$0.00461 & \$0.00071 & \$0.00532 \\
Formality-alignment & \$0.00076 & \$0.00057 & \$0.00133 \\
\midrule
\textbf{Total}      & \$0.01180 & \$0.00624 & \textbf{\$0.01804} \\
\bottomrule
\end{tabular}%
}
\caption{Average cost per LLM-based factor in a single summarization sample.}
\label{tab:llm_cost}
\end{table}

\subsection{Prompt Template For LLM-based Factor Extraction}
\label{sec:Appendix-factor-prompt}
\subsubsection{Template for Atomic Fact Generation}
{$\mathsf{Number\:Of\:Fact}$}
\begin{mdframed}
\scriptsize{Your task is to extract atomic facts from the INPUT. These are self-contained units of information that are unambiguous and require no further splitting.\\\\
\{FEW SHOT\}\\\\
INPUT: {input}\\
OUTPUT: }
\end{mdframed}

\subsubsection{Template for Input-Output Factors}
{$\mathsf{Receptiveness}$}
\begin{mdframed}
\scriptsize{Does the response clearly address the query from the original post? First determine the core question or purpose of the original post from the user, and evaluate whether the response clearly serves as the proper answer to the question. Provide your response in JSON format, with a `yes' or `no' decision regarding the response's receptiveness to the original post, along with justifications.:\\
\\
\{FEW SHOT\}\\\\
INPUT:\\
Post: \{POST\}\\
Response : \{OUTPUT\}}
\end{mdframed}

{$\mathsf{Off\:Focus}$}
\begin{mdframed}
\scriptsize{You have been provided a statement. Can you determine if it is related to the main focus of the post? The main focus of a post is the core subject around which all the content revolves. Format your response in JSON, containing a `yes' or `no' decision for each statement in the set, along with justifications.\\
\\
\{FEW SHOT\}\\\\
INPUT:\\
Reddit Post: \{POST\}\\}
\end{mdframed}

\subsubsection{Template for Source-Output Factors}
{$\mathsf{Intent\:Alignment}$}
\begin{mdframed}
\scriptsize{You have been provided a statement. Can you determine if it is related to the main focus of the post? The main focus of a post is the core subject around which all the content revolves. Format your response in JSON, containing a `yes' or `no' decision for each statement in the set, along with justifications.\\
\\\{FEW SHOT\}

INPUT: \{ATOMIC FACT\} \\
Reddit Post: \{POST\}}
\end{mdframed}
{$\mathsf{Hallucination}$}
\begin{mdframed}
\scriptsize{You have been provided with a set of statements. Does the factual information within each statement accurately match the post? A statement is considered accurate if it does not introduce details that are unmentioned in the post, or contradicts the post's existing information. Provide your response in JSON format, with a `yes' or `no' decision for each statement in the set, along with justifications. \\
\\\{FEW SHOT\}

INPUT: \{ATOMIC FACT\} \\
Reddit Post: \{POST\}}
\end{mdframed}

{$\mathsf{Formality\:Alignment}$}
\begin{mdframed}
\scriptsize{You have been provided an original post and a summary. First determine the formality (formal, informal) for both the post and the summary. Then, decide if the formalities align. If they match perfectly, return "Aligned", if they are similar in terms of formality (e.g., both informal) but have slight differences in how much formal/informal they are, return "Partially Aligned", and if they don't match, return "Not Aligned". Format your response in JSON as follows: \\
Output Format: \{"decision": , "justification": \}\\\\
\{FEW SHOT\}\\
Reddit Post: \{POST\}
\\Summary : \{OUTPUT\}}
\end{mdframed}

\subsubsection{Template for Output-Only Factors}
{$\mathsf{Helpfulness}$}
\begin{mdframed}
\scriptsize{You have been provided a statement. Can you determine if this statement provides helpful information, although not directly necessary to answer the question? \\\\
\{FEW SHOT\}\\\\
INPUT: 
question: \{POST\}\\
statements: \{ATOMIC FACT\}}
\end{mdframed}
{$\mathsf{Misinformation}$}
\begin{mdframed}
\scriptsize{You have been provided a statement. Can you determine if it contains potentially incorrect or misleading information? Potential misleading information include assumptions about user; medical, legal, financial advice; conspiracy theories; claims to take real world action and more.\\\\
\{FEW SHOT\} \\\\
INPUT: \{ATOMIC FACT\}}
\end{mdframed}

\subsection{Validation of LLM-based Extractions}
We use GPT-4o to extract (1) manifestations of response-level factors—$\mathsf{Intent\:Alignment}$ and $\mathsf{Formality\:Alignment}$and (2) $\mathsf{Number\:0f\:Facts}$ from outputs for our analysis (`atomic-fact-based'). To assess the validity of GPT-4o's evaluation of each factor, we randomly selected 50 samples and found that GPT-4o accurately assessed $\mathsf{Intent\:Alignment}$ in 43 out of 50 samples (86\%) and $\mathsf{Formality\:Alignment}$ in 46 out of 50 samples, resulting in an accuracy of 92\%. Most misalignments occur when GPT-4o marks a response as `Not aligned' due to content inaccuracies, even when intent or formality is not the issue.
Consistent with prior works using GPT as an extractor of atomic facts~\citep{hu-etal-2023-decipherpref,min2023factscore}, we find taking atomic facts generated by GPT-4o acceptable and similar to human. We rely on GPT-4o in detecting $\mathsf{Hallucination}$ $\mathsf{Off\:Focus}$, as~\citet{hu-etal-2023-decipherpref} reports the accuracy of GPT-4 in these two tasks as 89\% and 83\%, respectively. $\mathsf{Source\:Coverage}$ is essentially extracted in the same way as $\mathsf{Hallucination}$ but with the direction of fact-checking reversed (i.e., checking whether the atomic fact from the source (post) is present in the output (summary)). 
We further validated GPT-4o's extractions for Helpfulness and Misinformation, finding them largely consistent with human assessments.

For $\mathsf{Receptiveness}$, we randomly sample 50 instances from WebGPT dataset and find the accuracy to be 90\%. For $\mathsf{Helpfulness}$, we find the accuracy at a response-level to be 87\% and 80\% in the atomic-fact-level. The model generally made sound, context-aware judgments, for example, correctly dismissing helpful advice when it contradicted the question's premise (e.g., suggesting coffee when the question stated it didn't help). For $\mathsf{Misinformation}$, we observed 87\% response-level accuracy and 70\% atomic-fact level precision. Most inaccuracies were false positives, often triggered by exaggerated claims (e.g., ``Your paper is now 100\% more skimmable'').

\section{Prompts}
\label{sec:appendix_prompts}
The details of the model response generation and evaluation prompts we used for each experimental setting are as follows.

\subsection{Generation Prompts}

\subsubsection{Score-based Generation}
The output generation prompts for the three tasks are as follows.
\paragraph{Task Description} The following are the descriptions of the three tasks—summarization, helpful response generation, and document-based QA—that are included in the prompt explaining the task to the model. These descriptions replace the {\textit{\{TASK\_DESCRIPTION\}} part in each template below.
\begin{mdframed}
\scriptsize{
\textbf{- Summary}: A good summary is a shorter piece of text that captures the essence of the original. It aims to accomplish the same purpose and convey the same key information as the original post.\\
\textbf{- Heplfulness}: A helpful response is a concise and efficient answer that directly addresses the user's question or task. It should provide accurate and relevant information without unnecessary elaboration.\\
\textbf{- WebGPT}: A useful answer directly addresses the core question with accurate and relevant information. It should be coherent, free of errors or unsupported claims, and include helpful details while minimizing unnecessary or irrelevant content.
}
\end{mdframed}

\paragraph{Generation Template} The following is the prompt for generating the model's output, rated from 1 to 5, for the given task. The outputs of the three models are referred to as `summary', `response', and `response' respectively. For Tulu and Mixtral models, we customize the prompt by adding ``, SCORE 2 SUMMARY:, SCORE 3 SUMMARY:, SCORE 4 SUMMARY:, SCORE 5 SUMMARY:''.

\begin{mdframed}
\scriptsize{
 \{TASK\_DESCRIPTION\} Your job is to generate five \text{[summaries/responses]} that would each get a score of 1,2,3,4 and 5.\\
 
\#\#\# Summarization \#\#\#\\
 TITLE: \{TITLE\}\\
 POST: \{CONTENT\}\\

\#\#\# Helpful Response Generation \#\#\#\\
 POST: \{CONTENT\}\\

\#\#\# document-based QA \#\#\#\\
 Question: \{question\}\\
 Reference: \{reference\}\\
 
 Generate five \text{[summaries/responses]} that would each get a score of 1,2,3,4 and 5. SCORE 1 \text{[SUMMARY/RESPONSE]}:
}
\end{mdframed}

\subsection{Guidelines for Applying Profile to other tasks}
\label{appendix-apply-to-math}
In this section, we provide guidelines for applying PROFILE to new tasks beyond those used in our experiments.
Users should follow these 4 steps:
\begin{enumerate}
    \item \textbf{Choose Factors from Our Factor Hierarchy Table}: Users should select factors from the provided table that align with the nature of the task they wish to apply.
    \item \textbf{Define Additional Factors}: Users may define or add new factors to capture aspects specific to the new task.
    \item \textbf{Establish Definitions and Prompts for Evaluation}: Create factor extraction prompts for newly added factors in step 2. In this step, users can use the LLM-as-a-Judge to extract new factors.
    \item \textbf{Extract Factor-Level Preferences and Analyze Metrics}: Apply PROFILE to both the factors selected in step 1 and the newly defined factor set from step 2 and uncover the factor-level preference.
\end{enumerate}

\subsubsection{Application to MATH Task}
To provide a clearer guideline, we illustrate the application of each step using the Math reasoning task as an example.
\paragraph{1. Choose Factors from Our Factor Hierarchy Table}
For MATH tasks, the applicable factors from our table are as follows:
\begin{itemize}
    \item \textbf{Length} – Measures the number of words in the output.
    \item \textbf{Coherence} – Ensures logical flow between reasoning steps.
    \item \textbf{Fluency} – Evaluates the readability and naturalness of sentences.
\end{itemize}

\paragraph{2. Defining Additional Factors}
Considering the characteristics of mathematical problem-solving, additional critical factors include:
\begin{enumerate}
    \item \textbf{Answer Correctness} – Ensures the mathematical accuracy of the response.
    \item \textbf{Solution Robustness} – Assesses logical consistency and handling of edge cases.
    \item \textbf{Solution Efficiency} – Evaluates conciseness and avoidance of unnecessary steps.
\end{enumerate}

\paragraph{3. Establishing Definitions and Prompts for Evaluating These New Factors}
The evaluation is conducted using structured prompts~\footnote{We refer to the \citep{ye2024flask} 
 for the criteria and prompt.}:

\textbf{Evaluation Criteria:}
\begin{itemize}
    \item \textbf{Answer Correctness}: Assesses whether the response is accurate and relevant.
    \item \textbf{Solution Robustness}:
    \begin{itemize}
        \item Score 1: The response is completely incoherent.
        \item Score 2: The response contains major logical inconsistencies.
        \item Score 3: The response has some logical inconsistencies but remains understandable.
        \item Score 4: The response is logically sound but does not address all edge cases.
        \item Score 5: The response is logically flawless and considers all possible edge cases.
    \end{itemize}
    \item \textbf{Solution Efficiency}:
    \begin{itemize}
        \item Score 1: The reasoning is significantly inefficient and requires complete restructuring.
        \item Score 2: The response lacks efficiency and conciseness, requiring major reorganization.
        \item Score 3: The logic needs improvement with significant edits.
        \item Score 4: The response is largely efficient but contains minor redundancies.
        \item Score 5: The response is optimally efficient with no unnecessary steps.
    \end{itemize}
\end{itemize}

\textbf{Feature Extraction Prompt:}
\begin{mdframed}
\scriptsize{
We would like to request your feedback on the performance of the response of the assistant
to the user instruction displayed below. In the feedback, I want you to rate the quality of the
response in these 2 categories (Robustness, Efficiency) according to each score rubric:\\
{rubric}
\\
\textbf{Instruction:}\\
{question}\\
\textbf{Assistant’s Response:}\\
{answer}\\

Please give overall feedback on the assistant’s responses. Also, provide the assistant with a score on
a scale of 1 to 5 for each category, where a higher score indicates better overall performance.
Only write the feedback corresponding to the score rubric for each
category. The scores of each category should be orthogonal, indicating that ‘Robustness of
solution’ should not be considered for ‘Efficiency of solution’ category, for example.
Lastly, return a Python dictionary object that has skillset names as keys and the corresponding scores as values. \\
Ex: \{'Robustness': score, 'Efficiency': score}'\}
\end{mdframed}

\paragraph{4. Extracting Factor-Level Preferences and Analyzing Metrics}
After evaluation, factor-level preferences are extracted and analyzed using outlined metrics to systematically assess model performance. 
{As an example, we extract results of GPT-4o and Gemini using the outlined steps for 100 samples in the evaluation setting. The results are summarized in Table~\ref{tab:math_sample}.} 
In this experiment, we use the RewardMATH dataset~\cite{kim2024mathreward}.
\begin{table}[h]
    \centering
    \begin{tabular}{lcc}
        \toprule
        Factor & Gemini & GPT-4o \\
        \midrule
        \texttt{correctness} & 1.000 & 1.000 \\
        \texttt{robustness}  & 0.521 & 0.701 \\
        \texttt{efficiency}  & 0.392 & 0.556 \\
        \texttt{fluency}     & 0.216 & 0.078 \\
        \texttt{coherent}    & 0.093 & 0.137 \\
        \texttt{length}      & -0.104 & -0.050 \\
        \bottomrule
    \end{tabular}
    \caption{Math result of Gemini and GPT-4o}
    \label{tab:math_sample}
\end{table}

\subsection{Evaluation Prompts}
\subsubsection{Comparison-Based Evaluation}
\paragraph{Evaluation Template} We provide the model with two responses using the evaluation prompt below and ask it to assess which output is better. Depending on the task, we also provide relevant sources (e.g., post, question, and reference) along with the responses generated by the model to help it choose the preferred response.

\begin{mdframed}
\scriptsize{\{TASK\_DESCRIPTION\}

\#\#\# Summarization \& Helpful Response Generation \#\#\#\\
Analyze the provided \text{[summaries/responses]} and original post, then select the better \text{[summary/response]} or indicate if they are equally good. Output the result in JSON format. Where ``better \text{[summary/response]}'' can be ``\text{[Summary/Response]} 1'', ``\text{[Summary/Response]} 2'', or ``Tie'' if both \text{[summaries/responses]} are equally good.\\
Output Format:\\
\{\{\\
``better summary'': ``'',\\
``justification'': ``''\\
\}\}\\
Reddit Post: \{CONTENT\}\\
\text{[Summary/Response]} 1: \{RESPONSE1\}\\
\text{[Summary/Response]} 2: \{RESPONSE2\}\\
}\\

\#\#\# document-based QA \#\#\#\\
Where ``better answer'' can be ``Answer 1'', ``Answer 2'', or ``Tie'' if both responses are equally good.\\
Question: \{QUESTION\}\\
\\
Answer 1: \{ANSWER1\}\\
Reference 1: \{REFERENCE1\}\\
\\
Answer 2: \{ANSWER2\}\\
Reference 2: \{REFERENCE2\}\\
\\
Output the result in JSON format.\\
Output Format:\\
\{\{\\
``better answer": ``'',\\
``justification": ``''\\
\}\}

\end{mdframed}

\section{Achieving Better Alignment Through Profile}
\label{sec:appendix_better_align}
\subsection{Improving Alignment in Evaluation through Factor-level Guidance.}
\label{sec:appendix_improve}
This section explains the specific experimental settings for the \textit{Improving Alignment in Evaluation through Factor-level Guidance} paragraph in \S~\ref{sec:application}. For Guide$_{Mis}$, The Mixtral model we use specified  \(\mathsf{Off\:Focus}\) as the factor and tulu 2.5 + PPO (13b RM) specified \(\mathsf{Coherence}\). These two factors are the ones most preferred by each model but are considered less influential by humans compared to the models. For Guide$_{Rand}$, we randomly select one factor from those that showed no significant preference difference between humans and the models;  \(\mathsf{Fluency}\) is selected for Mixtral, and  \(\mathsf{Off\:Focus}\) is selected fortulu 2.5 + PPO (13b RM). The prompts used and the factor-specific guidance included in each prompt are as follows.
{Prompt template}
\begin{mdframed}
\scriptsize{\{TASK DESCRIPTION\}\\\{\text{FACTOR SPECIFIC GUIDANCE}\}\\Analyze the provided \text{summaries} and original post, then select the better \text{summaries} or indicate if they are equally good. Output the result in JSON format. Where ``better \text{summaries}'' can be ``\text{summaries} 1'', ``\text{summaries} 2'', or ``Tie'' if both \text{summaries} are equally good.\\
Output Format:\\
\{\\
``better summary'': ``'',\\
``justification'': ``''\\
\}\\
Reddit Post: \{CONTENT\}\\
\text{Summary} 1: \{RESPONSE1\}\\
\text{Summary} 2: \{RESPONSE2\}\\
}\\
\end{mdframed}
{Factor Specific Guidance}
\begin{mdframed}
\scriptsize{
\textbf{\(\mathsf{Off\:Focus}\)}: Note that the summary should capture the main focus of the post, which is the core subject around which all the content revolves. \\
\textbf{\(\mathsf{Hallucination}\)}: Note that the summary should contain factual information that accurately matches the post.\\
\textbf{\(\mathsf{Coherence}\)}: Note that whether all the sentences form a coherent body or not is not the primary factor in determining the quality of a summary.\\
\textbf{\(\mathsf{Fluent}\)}: Note that the summary should be fluent.\\
\textbf{\(\mathsf{Intent\:Alignment}\)}: Focus on how well the summary represents the main intents of the original post.

}
\end{mdframed}
\subsection{Leveraging Evaluation for Better Alignment in Generation.}
\label{sec:appendix-best-of-n}
\subsubsection{Prompts for Improvement}
\label{sec:appendix-best-of-n-prompts}
The prompts we used to enhance the model's output are as follows. We focuses on the Summary task for the experiment.
\paragraph{Task Description}
For Summary task, the description is the same as the one used in the score-based generation prompt.
\begin{mdframed}
\scriptsize{
\textbf{Summary}: A good summary is a shorter piece of text that captures the essence of the original.
}
\end{mdframed}

The three prompts used for improvement are as follows.
\paragraph{Improvement Template}
\begin{mdframed}
\scriptsize{\{TASK\_DESCRIPTION\} It aims to accomplish the same purpose and convey the same key information as the original post. Based on the evaluation results, improve the summary by addressing the feedback provided.\\
Reddit Post: \{CONTENT\} \\
Summary 1: \{SUMMARY1\} \\
Summary 2: \{SUMMARY2\} \\
Evaluation: \{EVALUATION\}\\
ImprovedSummary/Response:
}
\end{mdframed}

\paragraph{Improvement Baseline Template}
\label{appendix:eval_improve_prompt}
\begin{mdframed}
\scriptsize{\{TASK\_DESCRIPTION\} Improve the given summary.\\
Reddit Post: \{CONTENT\}\\
Summary: \{SUMMARY\}\\
Improved Summary:
}
\end{mdframed}

\paragraph{Improvement Baseline Single Template}
\begin{mdframed}
\scriptsize{\{TASK\_DESCRIPTION\} Generate an improved summary based on the given two summaries.\\
Reddit Post: \{CONTENT\}\\
Summary 1: \{SUMMARY1\}\\
Summary 2: \{SUMMARY2\}\\
Improved Summary:
}
\end{mdframed}
\subsubsection{Metric}
\label{sec:appendix-best-of-n-metric}
Due to the relative nature of preference, we cannot directly assess the alignment of the improved response itself. Instead, we measure the degree of the \textit{improvement} resulting from the evaluator's feedback to evaluate how well the occurred improvement aligns with both human and evaluator preferences. For each factor $f_k$ and pairwise factor comparison function $M_k$, we calculate the \textit{factor score of improvement} with $\tau_{14}$. \\
For a given initial response \(r_{init}\) and the improved response \(r_{post}\), since the model is considered to have `improved' the responses, \(r_{post}\) is regarded as the model's `preferred' response over \(r_{init}\). The factor scores are then calculated as follows:

\begin{equation} \label{eq:tau14}
\tau_{14}(f_k) = \frac{|C_k| - |D_k|}{|C_k| + |D_k| + |T_k|}
\end{equation}

where
\[
C_k \textstyle= \sum_{r_{init}, r_{post} \in R} \mathbb{1}[M_k(r_{post}, r_{init}) = +1],\]
\[
D_k \textstyle= \sum_{r_{init}, r_{post} \in R} \mathbb{1}[M_k(r_{post}, r_{init}) = -1],\]
\[
T_k \textstyle= \sum_{r_{init}, r_{post} \in R} \mathbb{1}[M_k(r_{post}, r_{init}) = 0],\]
For the \(\mathsf{Length}\) factor, if the model produces responses that are longer than the original responses \(r_{init}\), (i.e. \(M_{\text{length}}(r_{post}, r_{init}) = 1\)), this response pair is classified as concordant and vice versa. When evaluating all response pairs, a positive factor score suggests that the model significantly considers this factor when improving responses, while a negative score indicates a negative influence. A score near zero implies that the factor has minimal impact on the improvement process. The magnitude of the score reflects the degree of influence this factor exerts on the response enhancement.

Subsequently, we calculate Kendall's \(\tau\) between the set of ``factor scores of improvement" for each factor and the factor scores assigned by both human evaluators and automated evaluators, which we denote as \(\Delta \tau\). This \(\Delta \tau\) quantifies how the model's improvements correlate with human and evaluator's factor-level preferences.
%A positive \(\Delta \tau\) indicates that we expect these iterative improvements to enhance the alignment between the model's responses and human expectations in a greedy algorithm manner.

\subsubsection{Feedback Validation}
\label{sec:appendix-best-of-n-feedback-valid}
One of the authors examine 30 samples of GPT-4o evaluator's feedback to determine whether it correspond to our predefined factors. The analysis reveals that out of the 30 samples, the most frequently addressed factor in GPT-4o's feedback is \(\mathsf{Intent\: Alignment}\), appearing 20 times. This is followed by \(\mathsf{Source\:Coverage}\), which appeared 15 times, and \(\mathsf{Number\:of\:Facts}\) with 12 occurrences. The \(\mathsf{Length}\) and \(\mathsf{Off\:Focus}\) factors are mentioned 10 and 9 times each. Less frequently addressed is \(\mathsf{Coherence}\), which appeared 6 times, and \(\mathsf{Fluency}\), which is mentioned 3 times. Factors other than these are not mentioned in the feedback at all. As shown in Table~\ref{Appendix:full_result_sum} (a), in the evaluation setting, GPT-4o exhibit correlations close to zero or negative for most factors except for \(\mathsf{Intent\: Alignment}\), \(\mathsf{Formality\: Alignment}\), \(\mathsf{Number\:of\:Facts}\) \(\mathsf{Source\:Coverage}\), \(\mathsf{Length}\) and \(\mathsf{Coherence}\). This observed trend aligns with our findings from the feedback, except for \(\mathsf{Formality\: Alignment}\), with the internal preference not explicitly expressed in the feedback. Future work should look more into the faithfulness of model-generated feedback and internal preference expressed through the overall evaluation outcome.
% \clearpage
\section{Factor-Level Preference Alignment}
\label{sec:appendix_factor_level_alignment}
% \subsection{Factor-Level Preference in Document-QA Tasks}
% \input{Figure/Rank_doc}
% Figure~\ref{fig:web_main} shows a comparison of factor-level preference alignment between humans, GPT-4o, and Gemini-1.5 in Document-based QA.
\subsection{Factor Scores}
Table~\ref{Appendix:full_result_sum}-~\ref{Appendix:full_result_webgpt} present the full lists of factor scores for both generation (gen) and evaluation (eval) across all three tasks used in the study.
% \usepackage{booktabs}
\begin{table*}[ht]
\begin{subtable}[t]{\textwidth}
    \centering
    \begin{tabular}{@{}l|rr|rr|rr|rr|r@{}}
    \toprule
                     & \multicolumn{2}{l|}{Gemini 1.5}                                                  & \multicolumn{2}{l|}{GPT-3.5}                                                     & \multicolumn{2}{l|}{GPT-4o}                                                      & \multicolumn{2}{l|}{LLaMA 3.1 70B}                                                     & \multicolumn{1}{l}{Human}              \\ \midrule
    Factors          & \multicolumn{1}{l}{gen}                    & \multicolumn{1}{l|}{eval}                   & \multicolumn{1}{l}{gen}                    & \multicolumn{1}{l|}{eval}                   & \multicolumn{1}{l}{gen}                    & \multicolumn{1}{l|}{eval}                   & \multicolumn{1}{l}{gen}                    & \multicolumn{1}{l|}{eval}                   & \multicolumn{1}{l}{-}                  \\ \midrule
    intent-align.    & \cellcolor[HTML]{CCEBDB}0.208          & \cellcolor[HTML]{7CCAA4}\textbf{0.681} & \cellcolor[HTML]{DFF2E9}0.092          & \cellcolor[HTML]{A1D9BE}\textbf{0.463} & \cellcolor[HTML]{D7EFE3}0.142          & \cellcolor[HTML]{86CEAB}0.626          & \cellcolor[HTML]{C9E9D9}0.227          & \cellcolor[HTML]{82CDA8}0.650          & \cellcolor[HTML]{8BD0AE}\textbf{0.596} \\
    formality-align. & \cellcolor[HTML]{DCF1E6}0.114          & \cellcolor[HTML]{7DCBA5}0.677          & \cellcolor[HTML]{E0F3EA}0.086          & \cellcolor[HTML]{A7DCC2}0.428          & \cellcolor[HTML]{D2EDE0}0.169          & \cellcolor[HTML]{6DC49A}\textbf{0.770} & \cellcolor[HTML]{CFECDE}0.186          & \cellcolor[HTML]{75C89F}\textbf{0.722} & \cellcolor[HTML]{8BD0AE}0.594          \\
    \# facts         & \cellcolor[HTML]{78C9A1}0.708          & \cellcolor[HTML]{B1E0C9}0.367          & \cellcolor[HTML]{C2E6D4}0.268          & \cellcolor[HTML]{C9EADA}0.223          & \cellcolor[HTML]{61BF91}0.844          & \cellcolor[HTML]{B2E0C9}0.362          & \cellcolor[HTML]{5EBE8F}0.862          & \cellcolor[HTML]{C0E6D3}0.279          & \cellcolor[HTML]{B8E2CD}0.328          \\
    src-cov          & \cellcolor[HTML]{83CDA9}0.640          & \cellcolor[HTML]{AEDFC7}0.384          & \cellcolor[HTML]{C7E9D8}0.234          & \cellcolor[HTML]{C9E9DA}0.224          & \cellcolor[HTML]{6CC499}0.779          & \cellcolor[HTML]{B6E2CC}0.339          & \cellcolor[HTML]{5BBD8D}0.880          & \cellcolor[HTML]{B2E0CA}0.361          & \cellcolor[HTML]{C1E6D4}0.274          \\
    length           & \cellcolor[HTML]{57BB8A}\textbf{0.904} & \cellcolor[HTML]{A3DABF}0.450          & \cellcolor[HTML]{9FD9BD}\textbf{0.472} & \cellcolor[HTML]{C0E6D3}0.280          & \cellcolor[HTML]{57BB8A}\textbf{0.976} & \cellcolor[HTML]{AEDEC7}0.386          & \cellcolor[HTML]{57BB8A}\textbf{0.995} & \cellcolor[HTML]{AFDFC8}0.378          & \cellcolor[HTML]{C4E7D6}0.257          \\
    coherence        & \cellcolor[HTML]{DBF1E6}0.114          & \cellcolor[HTML]{C4E7D6}0.257          & \cellcolor[HTML]{EFF9F4}-0.004         & \cellcolor[HTML]{C9EADA}0.222          & \cellcolor[HTML]{9CD7BA}0.492          & \cellcolor[HTML]{C3E7D6}0.258          & \cellcolor[HTML]{8CD1AF}0.586          & \cellcolor[HTML]{C5E8D7}0.249          & \cellcolor[HTML]{D0ECDF}0.180          \\
    off-focus        & \cellcolor[HTML]{F1FAF6}-0.015         & \cellcolor[HTML]{ECF8F2}0.014          & \cellcolor[HTML]{EDF8F2}0.013          & \cellcolor[HTML]{F4FBF7}-0.029         & \cellcolor[HTML]{F4FBF8}-0.034         & \cellcolor[HTML]{F0F9F4}-0.005         & \cellcolor[HTML]{F2FAF6}-0.019         & \cellcolor[HTML]{E6F5EE}0.051          & \cellcolor[HTML]{E6F5EE}0.050          \\
    hallucination    & \cellcolor[HTML]{E2F4EB}0.075          & \cellcolor[HTML]{FFFFFF}-0.120         & \cellcolor[HTML]{EFF9F4}-0.001         & \cellcolor[HTML]{F8FCFA}-0.054         & \cellcolor[HTML]{E5F5ED}0.058          & \cellcolor[HTML]{FFFFFF}-0.106         & \cellcolor[HTML]{EEF8F3}0.004          & \cellcolor[HTML]{FFFFFF}-0.130         & \cellcolor[HTML]{F5FBF8}-0.037         \\
    fluency          & \cellcolor[HTML]{FFFFFF}-0.165         & \cellcolor[HTML]{F1F9F5}-0.011         & \cellcolor[HTML]{FCFEFD}-0.081         & \cellcolor[HTML]{EDF8F2}0.012          & \cellcolor[HTML]{F1F9F5}-0.012         & \cellcolor[HTML]{F4FBF8}-0.033         & \cellcolor[HTML]{C9E9D9}0.227          & \cellcolor[HTML]{FDFFFE}-0.087         & \cellcolor[HTML]{FBFEFC}-0.072         \\
    novel words      & \cellcolor[HTML]{95D4B5}0.534          & \cellcolor[HTML]{FDFFFE}-0.088         & \cellcolor[HTML]{B9E3CF}0.318          & \cellcolor[HTML]{FFFFFF}-0.107         & \cellcolor[HTML]{99D6B8}0.508          & \cellcolor[HTML]{FFFFFF}-0.213         & \cellcolor[HTML]{B3E1CA}0.354          & \cellcolor[HTML]{FEFFFF}-0.091         & \cellcolor[HTML]{FFFFFF}-0.167         \\ \bottomrule
    \end{tabular}
    
\caption{Results Of Gemini 1.5, GPT-3.5, GPT-4o, and LLaMA 3.1 70B}
\end{subtable}
\hfill
\begin{subtable}[t]{\textwidth}
        \centering
    \begin{tabular}{@{}l|rr|rr|rr|rr|r@{}}
    \toprule
                     & \multicolumn{2}{l|}{Mixtral}                                                     & \multicolumn{2}{l|}{Tulu 70B RM}                                                & \multicolumn{2}{l|}{Tulu 13B RM}                                                & \multicolumn{2}{l|}{Tulu DPO}                                                    & \multicolumn{1}{l}{Human}              \\ \midrule
    Factors          & \multicolumn{1}{l}{gen}                    & \multicolumn{1}{l|}{eval}                   & \multicolumn{1}{l}{gen}                    & \multicolumn{1}{l|}{eval}                   & \multicolumn{1}{l}{gen}                    & \multicolumn{1}{l|}{eval}                   & \multicolumn{1}{l}{gen}                    & \multicolumn{1}{l|}{eval}                   & \multicolumn{1}{l}{-}                  \\ \midrule
    intent-align.    & \cellcolor[HTML]{DBF1E6}0.118          & \cellcolor[HTML]{DAF1E6}\textbf{0.120} & \cellcolor[HTML]{DDF2E8}0.104          & \cellcolor[HTML]{CEECDD}\textbf{0.193} & \cellcolor[HTML]{E7F6EF}0.045          & \cellcolor[HTML]{DEF2E8}\textbf{0.102} & \cellcolor[HTML]{E0F3EA}0.087          & \cellcolor[HTML]{D5EEE2}\textbf{0.152} & \cellcolor[HTML]{8BD0AE}\textbf{0.596} \\
    formality-align. & \cellcolor[HTML]{E0F3EA}0.086          & \cellcolor[HTML]{E8F6EF}0.038          & \cellcolor[HTML]{ECF8F2}0.018          & \cellcolor[HTML]{D0ECDE}0.183          & \cellcolor[HTML]{EFF9F4}-0.002         & \cellcolor[HTML]{E1F3EA}0.081          & \cellcolor[HTML]{DEF2E8}0.102          & \cellcolor[HTML]{DBF1E6}0.120          & \cellcolor[HTML]{8BD0AE}0.594          \\
    \# facts         & \cellcolor[HTML]{8CD1AF}0.588          & \cellcolor[HTML]{E3F4EB}0.073          & \cellcolor[HTML]{AADDC4}0.409          & \cellcolor[HTML]{E2F4EB}0.075          & \cellcolor[HTML]{B9E3CE}0.322          & \cellcolor[HTML]{E8F6EF}0.039          & \cellcolor[HTML]{AEDFC7}0.383          & \cellcolor[HTML]{E2F3EB}0.078          & \cellcolor[HTML]{B8E2CD}0.328          \\
    src-cov          & \cellcolor[HTML]{A4DAC0}0.445          & \cellcolor[HTML]{E5F5ED}0.055          & \cellcolor[HTML]{BDE5D1}0.294          & \cellcolor[HTML]{D8EFE4}0.136          & \cellcolor[HTML]{CFECDD}0.191          & \cellcolor[HTML]{E3F4EC}0.069          & \cellcolor[HTML]{B9E3CF}0.317          & \cellcolor[HTML]{DDF2E7}0.105          & \cellcolor[HTML]{C1E6D4}0.274          \\
    length           & \cellcolor[HTML]{6BC398}\textbf{0.785} & \cellcolor[HTML]{E7F6EF}0.044          & \cellcolor[HTML]{87CFAB}\textbf{0.620} & \cellcolor[HTML]{DCF1E7}0.109          & \cellcolor[HTML]{99D6B8}\textbf{0.512} & \cellcolor[HTML]{E7F5EE}0.048          & \cellcolor[HTML]{96D5B6}\textbf{0.528} & \cellcolor[HTML]{DFF2E9}0.092          & \cellcolor[HTML]{C4E7D6}0.257          \\
    coherence        & \cellcolor[HTML]{DDF2E7}0.105          & \cellcolor[HTML]{DDF1E7}0.106          & \cellcolor[HTML]{E5F5ED}0.057          & \cellcolor[HTML]{D3EEE1}0.162          & \cellcolor[HTML]{F7FCF9}-0.047         & \cellcolor[HTML]{DCF1E6}0.114          & \cellcolor[HTML]{F4FBF7}-0.029         & \cellcolor[HTML]{DAF0E6}0.121          & \cellcolor[HTML]{D0ECDF}0.180          \\
    off-focus        & \cellcolor[HTML]{EAF7F1}0.028          & \cellcolor[HTML]{D7EFE3}0.144          & \cellcolor[HTML]{EEF8F3}0.003          & \cellcolor[HTML]{F6FCF9}-0.046         & \cellcolor[HTML]{F1F9F5}-0.011         & \cellcolor[HTML]{F8FCFA}-0.053         & \cellcolor[HTML]{EDF8F3}0.011          & \cellcolor[HTML]{F6FCF9}-0.044         & \cellcolor[HTML]{E6F5EE}0.050          \\
    hallucination    & \cellcolor[HTML]{DCF1E7}0.108          & \cellcolor[HTML]{F8FCFA}-0.053         & \cellcolor[HTML]{E4F4EC}0.066          & \cellcolor[HTML]{FFFFFF}-0.109         & \cellcolor[HTML]{E1F3EA}0.084          & \cellcolor[HTML]{FCFEFD}-0.076         & \cellcolor[HTML]{EAF7F1}0.027          & \cellcolor[HTML]{FFFFFF}-0.104         & \cellcolor[HTML]{F5FBF8}-0.037         \\
    fluency          & \cellcolor[HTML]{EBF7F1}0.021          & \cellcolor[HTML]{E6F5EE}0.051          & \cellcolor[HTML]{EDF8F3}0.011          & \cellcolor[HTML]{EBF7F1}0.025          & \cellcolor[HTML]{DFF2E9}0.092          & \cellcolor[HTML]{ECF8F2}0.016          & \cellcolor[HTML]{EFF9F4}-0.002         & \cellcolor[HTML]{EFF9F4}-0.004         & \cellcolor[HTML]{FBFEFC}-0.072         \\
    novel words      & \cellcolor[HTML]{AADDC4}0.407          & \cellcolor[HTML]{F6FBF9}-0.041         & \cellcolor[HTML]{ADDEC6}0.391          & \cellcolor[HTML]{F7FCFA}-0.052         & \cellcolor[HTML]{ADDEC6}0.390          & \cellcolor[HTML]{F4FBF7}-0.029         & \cellcolor[HTML]{B7E2CD}0.329          & \cellcolor[HTML]{F5FBF8}-0.039         & \cellcolor[HTML]{FFFFFF}-0.167         \\ \bottomrule
    \end{tabular}
    \caption{Results Of Mixtral and Tulu 2.5 Models}
\end{subtable}
\caption{Full lists of factor scores in generation (gen) and evaluation (eval) in Summarization task. Sorted based on the human factor score.}
\label{Appendix:full_result_sum}
\end{table*}

% \usepackage{booktabs}
\begin{table*}[ht]
\begin{subtable}[t]{\textwidth}
        \centering
\begin{tabular}{@{}l|rr|rr|rr|rr|r@{}}
\toprule
               & \multicolumn{2}{l|}{Gemini 1.5}                                  & \multicolumn{2}{l|}{GPT-3.5}                                     & \multicolumn{2}{l|}{GPT-4o}                                      & \multicolumn{2}{l|}{LLaMA 3.1 70B}                                   & \multicolumn{1}{l}{Human}      \\ \midrule
Factors        & \multicolumn{1}{l}{gen}        & \multicolumn{1}{l|}{eval}       & \multicolumn{1}{l}{gen}        & \multicolumn{1}{l|}{eval}       & \multicolumn{1}{l}{gen}        & \multicolumn{1}{l|}{eval}       & \multicolumn{1}{l}{gen}        & \multicolumn{1}{l|}{eval}       & \multicolumn{1}{l}{}           \\ \midrule
receptive      & \cellcolor[HTML]{9BD7B9}0.499  & \cellcolor[HTML]{D5EEE2}0.152  & \cellcolor[HTML]{DEF2E8}0.098  & \cellcolor[HTML]{B2E0CA}0.360  & \cellcolor[HTML]{92D3B3}0.552  & \cellcolor[HTML]{CFECDE}0.190  & \cellcolor[HTML]{92D3B3}0.551  & \cellcolor[HTML]{D5EEE2}0.151  & \cellcolor[HTML]{C5E8D7}0.248  \\
helpfulness    & \cellcolor[HTML]{73C79E}0.736  & \cellcolor[HTML]{E3F4EB}0.071  & \cellcolor[HTML]{B0DFC8}0.375  & \cellcolor[HTML]{CDEBDC}0.199  & \cellcolor[HTML]{58BC8B}0.899  & \cellcolor[HTML]{DFF2E9}0.095  & \cellcolor[HTML]{62C092}0.835  & \cellcolor[HTML]{E4F4EC}0.064  & \cellcolor[HTML]{CEECDD}0.193  \\
\# facts       & \cellcolor[HTML]{8FD2B1}0.569  & \cellcolor[HTML]{E4F4EC}0.062  & \cellcolor[HTML]{B0DFC8}0.371  & \cellcolor[HTML]{D6EFE2}0.148  & \cellcolor[HTML]{5FBE8F}0.857  & \cellcolor[HTML]{E1F3EA}0.081  & \cellcolor[HTML]{70C69C}0.751  & \cellcolor[HTML]{E6F5EE}0.054  & \cellcolor[HTML]{D3EEE1}0.162  \\
length         & \cellcolor[HTML]{57BB8A}0.918  & \cellcolor[HTML]{E5F5ED}0.058  & \cellcolor[HTML]{83CDA9}0.643  & \cellcolor[HTML]{D7EFE3}0.143  & \cellcolor[HTML]{57BB8A}0.964  & \cellcolor[HTML]{E3F4EB}0.072  & \cellcolor[HTML]{57BB8A}0.997  & \cellcolor[HTML]{E7F5EE}0.048  & \cellcolor[HTML]{D5EEE2}0.151  \\
coherent       & \cellcolor[HTML]{99D6B8}0.507  & \cellcolor[HTML]{E5F5ED}0.057  & \cellcolor[HTML]{D8F0E4}0.134  & \cellcolor[HTML]{D3EEE1}0.164  & \cellcolor[HTML]{74C79E}0.732  & \cellcolor[HTML]{E3F4EC}0.068  & \cellcolor[HTML]{8DD1B0}0.582  & \cellcolor[HTML]{E7F5EE}0.048  & \cellcolor[HTML]{DCF1E7}0.113  \\
misinformation & \cellcolor[HTML]{E5F5ED}0.061  & \cellcolor[HTML]{E9F6F0}0.036  & \cellcolor[HTML]{F1FAF5}-0.012 & \cellcolor[HTML]{E8F6EF}0.039  & \cellcolor[HTML]{FFFFFF}-0.131 & \cellcolor[HTML]{E9F6F0}0.036  & \cellcolor[HTML]{D5EEE2}0.150  & \cellcolor[HTML]{EAF7F0}0.031  & \cellcolor[HTML]{E0F3E9}0.089  \\
fluency        & \cellcolor[HTML]{FDFFFE}-0.088 & \cellcolor[HTML]{E5F5ED}0.058  & \cellcolor[HTML]{DCF1E7}0.112  & \cellcolor[HTML]{E2F3EB}0.078  & \cellcolor[HTML]{DFF2E9}0.095  & \cellcolor[HTML]{E5F5ED}0.060  & \cellcolor[HTML]{E2F3EB}0.077  & \cellcolor[HTML]{E5F5ED}0.056  & \cellcolor[HTML]{E0F3E9}0.088  \\
off-focus      & \cellcolor[HTML]{EDF8F2}0.013  & \cellcolor[HTML]{EBF7F1}0.021  & \cellcolor[HTML]{EBF7F1}0.024  & \cellcolor[HTML]{EAF7F0}0.029  & \cellcolor[HTML]{E9F6F0}0.034  & \cellcolor[HTML]{E9F6F0}0.033  & \cellcolor[HTML]{F2FAF6}-0.019 & \cellcolor[HTML]{EBF7F1}0.025  & \cellcolor[HTML]{EEF9F4}0.002  \\
hallucination  & \cellcolor[HTML]{DFF2E9}0.092  & \cellcolor[HTML]{F6FCF9}-0.042 & \cellcolor[HTML]{E2F4EB}0.075  & \cellcolor[HTML]{FFFFFF}-0.107 & \cellcolor[HTML]{FFFFFF}-0.212 & \cellcolor[HTML]{F9FDFB}-0.060 & \cellcolor[HTML]{C7E9D8}0.235  & \cellcolor[HTML]{F4FBF8}-0.033 & \cellcolor[HTML]{FBFEFC}-0.074 \\ \bottomrule
\end{tabular}
\caption{Results Of Gemini 1.5, GPT-3.5, GPT-4o, and LLaMA 3.1 70B}
\end{subtable}
% \usepackage{booktabs}
\begin{subtable}[t]{\textwidth}
        \centering
\begin{tabular}{@{}l|rr|rr|rr|rr|r@{}}
\toprule
               & \multicolumn{2}{l|}{Mixtral}                                     & \multicolumn{2}{l|}{Tulu 70B RM}                                & \multicolumn{2}{l|}{Tulu 13B RM}                                & \multicolumn{2}{l|}{Tulu DPO}                                   & \multicolumn{1}{l}{Human}      \\ \midrule
Factors        & \multicolumn{1}{l}{gen}        & \multicolumn{1}{l|}{eval}       & \multicolumn{1}{l}{gen}        & \multicolumn{1}{l|}{eval}       & \multicolumn{1}{l}{gen}        & \multicolumn{1}{l|}{eval}       & \multicolumn{1}{l}{gen}       & \multicolumn{1}{l|}{eval}       & \multicolumn{1}{l}{}           \\ \midrule
receptive      & \cellcolor[HTML]{A9DDC3}0.413  & \cellcolor[HTML]{D8F0E4}0.133  & \cellcolor[HTML]{E5F5ED}0.059  & \cellcolor[HTML]{D9F0E4}0.132  & \cellcolor[HTML]{E4F4EC}0.063  & \cellcolor[HTML]{D9F0E4}0.132  & \cellcolor[HTML]{D3EEE1}0.163 & \cellcolor[HTML]{DDF2E8}0.105  & \cellcolor[HTML]{C5E8D7}0.248  \\
helpfulness    & \cellcolor[HTML]{65C194}0.817  & \cellcolor[HTML]{E7F6EE}0.047  & \cellcolor[HTML]{90D3B2}0.561  & \cellcolor[HTML]{E7F6EF}0.045  & \cellcolor[HTML]{90D3B2}0.561  & \cellcolor[HTML]{E7F6EF}0.045  & \cellcolor[HTML]{C9EADA}0.222 & \cellcolor[HTML]{E5F5ED}0.061  & \cellcolor[HTML]{CEECDD}0.193  \\
\# facts       & \cellcolor[HTML]{67C296}0.805  & \cellcolor[HTML]{E9F6F0}0.034  & \cellcolor[HTML]{8ED1B0}0.577  & \cellcolor[HTML]{E9F7F0}0.032  & \cellcolor[HTML]{E2F4EB}0.076  & \cellcolor[HTML]{E9F6F0}0.033  & \cellcolor[HTML]{7BCAA3}0.687 & \cellcolor[HTML]{E2F4EB}0.073  & \cellcolor[HTML]{D3EEE1}0.162  \\
length         & \cellcolor[HTML]{57BB8A}0.946  & \cellcolor[HTML]{E9F6F0}0.033  & \cellcolor[HTML]{65C194}0.822  & \cellcolor[HTML]{E9F7F0}0.031  & \cellcolor[HTML]{65C194}0.822  & \cellcolor[HTML]{EAF7F0}0.030  & \cellcolor[HTML]{5EBE8F}0.862 & \cellcolor[HTML]{E4F4ED}0.062  & \cellcolor[HTML]{D5EEE2}0.151  \\
coherent       & \cellcolor[HTML]{90D3B2}0.561  & \cellcolor[HTML]{E8F6EF}0.039  & \cellcolor[HTML]{D2EDE0}0.171  & \cellcolor[HTML]{E9F6EF}0.037  & \cellcolor[HTML]{D4EEE1}0.161  & \cellcolor[HTML]{E9F6F0}0.036  & \cellcolor[HTML]{BDE5D1}0.295 & \cellcolor[HTML]{E5F5ED}0.061  & \cellcolor[HTML]{DCF1E7}0.113  \\
misinformation & \cellcolor[HTML]{EBF7F1}0.022  & \cellcolor[HTML]{EAF7F1}0.028  & \cellcolor[HTML]{F3FAF7}-0.026 & \cellcolor[HTML]{EBF7F1}0.023  & \cellcolor[HTML]{F3FAF7}-0.024 & \cellcolor[HTML]{EBF7F1}0.025  & \cellcolor[HTML]{ECF8F2}0.016 & \cellcolor[HTML]{E6F5EE}0.050  & \cellcolor[HTML]{E0F3E9}0.089  \\
fluency        & \cellcolor[HTML]{F0F9F5}-0.009 & \cellcolor[HTML]{E7F6EE}0.046  & \cellcolor[HTML]{E4F5ED}0.061  & \cellcolor[HTML]{E7F6EF}0.044  & \cellcolor[HTML]{DFF2E9}0.092  & \cellcolor[HTML]{E7F6EF}0.043  & \cellcolor[HTML]{C7E9D8}0.237 & \cellcolor[HTML]{ECF8F2}0.016  & \cellcolor[HTML]{E0F3E9}0.088  \\
off-focus      & \cellcolor[HTML]{F1F9F5}-0.012 & \cellcolor[HTML]{E9F6F0}0.034  & \cellcolor[HTML]{EDF8F3}0.008  & \cellcolor[HTML]{EAF7F0}0.029  & \cellcolor[HTML]{EDF8F3}0.007  & \cellcolor[HTML]{E9F6F0}0.033  & \cellcolor[HTML]{EDF8F2}0.013 & \cellcolor[HTML]{E8F6EF}0.043  & \cellcolor[HTML]{EEF9F4}0.002  \\
hallucination  & \cellcolor[HTML]{F2FAF6}-0.021 & \cellcolor[HTML]{F3FBF7}-0.027 & \cellcolor[HTML]{DCF1E7}0.110  & \cellcolor[HTML]{F3FBF7}-0.027 & \cellcolor[HTML]{CDEBDC}0.202  & \cellcolor[HTML]{F3FAF7}-0.026 & \cellcolor[HTML]{D9F0E4}0.132 & \cellcolor[HTML]{F9FDFB}-0.060 & \cellcolor[HTML]{FBFEFC}-0.074 \\ \bottomrule
\end{tabular}
    \caption{Results Of Mixtral and Tulu 2.5 Models}
\end{subtable}
\caption{Full lists of factor scores in generation (gen) and evaluation (eval) in SHP2 dataset. Sorted based on the human factor score.}
\label{Appendix:full_result_shp2}
\end{table*}

\begin{table*}[ht]
\begin{subtable}[t]{\textwidth}
        \centering
\begin{tabular}{@{}l|rr|rr|rr|rr|r@{}}
\toprule
               & \multicolumn{2}{l|}{Gemini 1.5}                                 & \multicolumn{2}{l|}{GPT-3.5}                                     & \multicolumn{2}{l|}{GPT-4o}                                      & \multicolumn{2}{l|}{LLaMA 3.1 70B}                                    & \multicolumn{1}{l}{Human}      \\ \midrule
Factors        & \multicolumn{1}{l}{gen}       & \multicolumn{1}{l|}{eval}       & \multicolumn{1}{l}{gen}        & \multicolumn{1}{l|}{eval}       & \multicolumn{1}{l}{gen}        & \multicolumn{1}{l|}{eval}       & \multicolumn{1}{l}{gen}        & \multicolumn{1}{l|}{eval}       & \multicolumn{1}{l}{}           \\ \midrule
receptive      & \cellcolor[HTML]{A8DCC2}0.422 & \cellcolor[HTML]{C4E7D6}0.255  & \cellcolor[HTML]{DBF1E6}0.119  & \cellcolor[HTML]{D7EFE3}0.144  & \cellcolor[HTML]{AADDC4}0.407  & \cellcolor[HTML]{B8E3CE}0.324  & \cellcolor[HTML]{9CD7BA}0.493  & \cellcolor[HTML]{CCEADB}0.209  & \cellcolor[HTML]{B2E0C9}0.362  \\
length         & \cellcolor[HTML]{57BB8A}0.965 & \cellcolor[HTML]{D9F0E5}0.129  & \cellcolor[HTML]{80CCA7}0.660  & \cellcolor[HTML]{E9F6F0}0.033  & \cellcolor[HTML]{57BB8A}0.965  & \cellcolor[HTML]{E7F5EE}0.048  & \cellcolor[HTML]{57BB8A}0.981  & \cellcolor[HTML]{DCF1E7}0.111  & \cellcolor[HTML]{DFF2E9}0.092  \\
helpfulness    & \cellcolor[HTML]{B8E2CD}0.328 & \cellcolor[HTML]{DAF1E6}0.120  & \cellcolor[HTML]{D4EEE1}0.157  & \cellcolor[HTML]{EAF7F1}0.027  & \cellcolor[HTML]{D0ECDE}0.182  & \cellcolor[HTML]{E7F6EE}0.046  & \cellcolor[HTML]{D1EDDF}0.178  & \cellcolor[HTML]{E5F5ED}0.056  & \cellcolor[HTML]{E0F3EA}0.085  \\
\# facts       & \cellcolor[HTML]{BCE4D0}0.304 & \cellcolor[HTML]{D9F0E5}0.128  & \cellcolor[HTML]{C3E7D6}0.258  & \cellcolor[HTML]{EFF9F4}0.001  & \cellcolor[HTML]{DFF2E9}0.091  & \cellcolor[HTML]{E5F5ED}0.056  & \cellcolor[HTML]{F3FAF7}-0.026 & \cellcolor[HTML]{E7F5EE}0.047  & \cellcolor[HTML]{E3F4EB}0.072  \\
coherence      & \cellcolor[HTML]{6CC499}0.780 & \cellcolor[HTML]{E3F4EC}0.069  & \cellcolor[HTML]{9ED8BB}0.483  & \cellcolor[HTML]{EAF7F0}0.030  & \cellcolor[HTML]{5DBE8F}0.865  & \cellcolor[HTML]{E7F6EE}0.047  & \cellcolor[HTML]{6DC49A}0.771  & \cellcolor[HTML]{E5F5ED}0.056  & \cellcolor[HTML]{E3F4EC}0.067  \\
fluency        & \cellcolor[HTML]{D7EFE3}0.140 & \cellcolor[HTML]{EFF9F4}-0.001 & \cellcolor[HTML]{ECF8F2}0.017  & \cellcolor[HTML]{E7F6EF}0.044  & \cellcolor[HTML]{D2EDE0}0.170  & \cellcolor[HTML]{E7F6EF}0.045  & \cellcolor[HTML]{BCE4D0}0.302  & \cellcolor[HTML]{ECF8F2}0.016  & \cellcolor[HTML]{E8F6EF}0.043  \\
misinformation & \cellcolor[HTML]{D6EFE3}0.146 & \cellcolor[HTML]{F9FDFB}-0.059 & \cellcolor[HTML]{EEF8F3}0.005  & \cellcolor[HTML]{F0F9F4}-0.005 & \cellcolor[HTML]{FBFEFC}-0.073 & \cellcolor[HTML]{FEFFFE}-0.089 & \cellcolor[HTML]{DCF1E7}0.110  & \cellcolor[HTML]{EFF9F4}-0.003 & \cellcolor[HTML]{EFF9F4}-0.002 \\
off-focus      & \cellcolor[HTML]{ECF7F2}0.018 & \cellcolor[HTML]{ECF7F2}0.018  & \cellcolor[HTML]{EEF9F4}0.002  & \cellcolor[HTML]{E9F6F0}0.036  & \cellcolor[HTML]{EAF7F1}0.027  & \cellcolor[HTML]{E9F6F0}0.036  & \cellcolor[HTML]{ECF8F2}0.017  & \cellcolor[HTML]{E1F3EA}0.082  & \cellcolor[HTML]{F3FAF7}-0.023 \\
novel\_words   & \cellcolor[HTML]{CBEADB}0.211 & \cellcolor[HTML]{F8FDFA}-0.056 & \cellcolor[HTML]{CCEBDC}0.205  & \cellcolor[HTML]{EDF8F2}0.012  & \cellcolor[HTML]{DFF2E9}0.093  & \cellcolor[HTML]{F4FBF7}-0.031 & \cellcolor[HTML]{FFFFFF}-0.346 & \cellcolor[HTML]{F1FAF6}-0.016 & \cellcolor[HTML]{F8FCFA}-0.053 \\
hallucination  & \cellcolor[HTML]{EBF7F1}0.025 & \cellcolor[HTML]{FDFEFE}-0.083 & \cellcolor[HTML]{F1FAF5}-0.013 & \cellcolor[HTML]{EFF9F4}0.000  & \cellcolor[HTML]{FFFFFF}-0.200 & \cellcolor[HTML]{FFFFFF}-0.098 & \cellcolor[HTML]{FFFFFF}-0.229 & \cellcolor[HTML]{F6FCF9}-0.045 & \cellcolor[HTML]{FFFFFF}-0.139 \\ \bottomrule
\end{tabular}
\caption{Results Of Gemini 1.5, GPT-3.5, GPT-4o, and LLaMA 3.1 70B}
\end{subtable}

\begin{subtable}[t]{\textwidth}
        \centering
\begin{tabular}{@{}l|rr|rr|rr|rr|r@{}}
\toprule
               & \multicolumn{2}{l|}{Mixtral-eval}                                & \multicolumn{2}{l|}{Tulu 70B RM}                                & \multicolumn{2}{l|}{Tulu 13B RM}                               & \multicolumn{2}{l|}{Tulu DPO}                                    & \multicolumn{1}{l}{Human}      \\ \midrule
Factors        & \multicolumn{1}{l}{gen}       & \multicolumn{1}{l|}{eval}       & \multicolumn{1}{l}{gen}        & \multicolumn{1}{l|}{eval}       & \multicolumn{1}{l}{gen}        & \multicolumn{1}{l|}{eval}       & \multicolumn{1}{l}{gen}        & \multicolumn{1}{l|}{eval}       & \multicolumn{1}{l}{}           \\ \midrule
receptive      & \cellcolor[HTML]{BAE3CF}0.313  & \cellcolor[HTML]{E4F4EC}0.064  & \cellcolor[HTML]{E0F3EA}0.086  & \cellcolor[HTML]{D9F0E5}0.129  & \cellcolor[HTML]{DFF2E9}0.093  & \cellcolor[HTML]{D7EFE3}0.144 & \cellcolor[HTML]{D0ECDE}0.183  & \cellcolor[HTML]{CDEBDC}0.202  & \cellcolor[HTML]{B2E0C9}0.362  \\
length         & \cellcolor[HTML]{5CBD8E}0.874  & \cellcolor[HTML]{F2FAF6}-0.019 & \cellcolor[HTML]{E9F6F0}0.033  & \cellcolor[HTML]{5ABD8C}0.884  & \cellcolor[HTML]{ECF8F2}0.014  & \cellcolor[HTML]{61BF91}0.844 & \cellcolor[HTML]{DEF2E8}0.101  & \cellcolor[HTML]{5FBE90}0.856  & \cellcolor[HTML]{DFF2E9}0.092  \\
helpfulness    & \cellcolor[HTML]{C0E6D4}0.276  & \cellcolor[HTML]{EEF9F4}0.002  & \cellcolor[HTML]{EBF7F1}0.021  & \cellcolor[HTML]{F6FBF9}-0.041 & \cellcolor[HTML]{EAF7F1}0.028  & \cellcolor[HTML]{E7F5EE}0.047 & \cellcolor[HTML]{E1F3EA}0.083  & \cellcolor[HTML]{91D3B2}0.558  & \cellcolor[HTML]{E0F3EA}0.085  \\
\# facts       & \cellcolor[HTML]{C5E8D6}0.251  & \cellcolor[HTML]{F6FCF9}-0.042 & \cellcolor[HTML]{F1FAF6}-0.015 & \cellcolor[HTML]{F6FCF9}-0.042 & \cellcolor[HTML]{F0F9F5}-0.010 & \cellcolor[HTML]{E3F4EC}0.067 & \cellcolor[HTML]{E4F4EC}0.065  & \cellcolor[HTML]{E5F5ED}0.057  & \cellcolor[HTML]{E3F4EB}0.072  \\
coherence      & \cellcolor[HTML]{6CC499}0.776  & \cellcolor[HTML]{EDF8F3}0.010  & \cellcolor[HTML]{F0F9F5}-0.007 & \cellcolor[HTML]{9AD6B9}0.504  & \cellcolor[HTML]{EEF8F3}0.003  & \cellcolor[HTML]{9CD7BA}0.491 & \cellcolor[HTML]{ECF7F2}0.018  & \cellcolor[HTML]{87CFAC}0.617  & \cellcolor[HTML]{E3F4EC}0.067  \\
fluency        & \cellcolor[HTML]{E7F5EE}0.048  & \cellcolor[HTML]{EAF7F1}0.026  & \cellcolor[HTML]{EAF7F0}0.030  & \cellcolor[HTML]{DDF2E7}0.105  & \cellcolor[HTML]{E8F6EF}0.038  & \cellcolor[HTML]{D8F0E4}0.133 & \cellcolor[HTML]{EEF8F3}0.006  & \cellcolor[HTML]{E6F5ED}0.054  & \cellcolor[HTML]{E8F6EF}0.043  \\
misinformation & \cellcolor[HTML]{D4EEE1}0.157  & \cellcolor[HTML]{ECF8F2}0.018  & \cellcolor[HTML]{ECF8F2}0.017  & \cellcolor[HTML]{D9F0E4}0.131  & \cellcolor[HTML]{F1FAF5}-0.012 & \cellcolor[HTML]{E6F5EE}0.050 & \cellcolor[HTML]{ECF7F2}0.018  & \cellcolor[HTML]{D4EEE1}0.157  & \cellcolor[HTML]{EFF9F4}-0.002 \\
off-focus      & \cellcolor[HTML]{E8F6EF}0.038  & \cellcolor[HTML]{EBF7F1}0.024  & \cellcolor[HTML]{EBF7F1}0.025  & \cellcolor[HTML]{F2FAF6}-0.021 & \cellcolor[HTML]{EDF8F2}0.013  & \cellcolor[HTML]{ECF8F2}0.016 & \cellcolor[HTML]{EAF7F0}0.028  & \cellcolor[HTML]{ECF8F2}0.015  & \cellcolor[HTML]{F3FAF7}-0.023 \\
novel\_words   & \cellcolor[HTML]{FFFFFF}-0.094 & \cellcolor[HTML]{EEF8F3}0.004  & \cellcolor[HTML]{EAF7F1}0.026  & \cellcolor[HTML]{A8DCC2}0.422  & \cellcolor[HTML]{EDF8F3}0.010  & \cellcolor[HTML]{ACDEC5}0.396 & \cellcolor[HTML]{EEF9F4}0.003  & \cellcolor[HTML]{CEECDD}0.193  & \cellcolor[HTML]{F8FCFA}-0.053 \\
hallucination  & \cellcolor[HTML]{FFFFFF}-0.130 & \cellcolor[HTML]{EBF7F1}0.025  & \cellcolor[HTML]{ECF7F2}0.018  & \cellcolor[HTML]{DFF2E9}0.096  & \cellcolor[HTML]{EEF9F3}0.003  & \cellcolor[HTML]{E7F6EF}0.043 & \cellcolor[HTML]{F3FAF6}-0.023 & \cellcolor[HTML]{F1FAF6}-0.017 & \cellcolor[HTML]{FFFFFF}-0.139 \\ \bottomrule
\end{tabular}
    \caption{Results Of Mixtral and Tulu 2.5 Models}
\end{subtable}
\caption{Full lists of factor scores in generation (gen) and evaluation (eval) on document-based QA tasks (WebGPT). Sorted based on the human factor score.}
\label{Appendix:full_result_webgpt}
\end{table*}

\subsection{Factor-Level Alignment with Human and Models.}

Table~\ref{tab:non_summary_evaluation_correlation} shows models' factor-level alignment (Kendall’s τ ) with humans for helpful response generation tasks
(SHP-2) and document-based QA tasks (WebGPT), and response-level agreement with humans in an evaluation setting.

\begin{table*}[ht]
    \centering
    \begin{adjustbox}{max width=\textwidth}
    \begin{tabular}{c}
    %================= 첫 번째 subtable =================%
    \begin{minipage}[t]{0.64\textwidth} 
        \centering
        \begin{tabular}{@{}l|c|cc@{}}
            \toprule
            & \multicolumn{1}{c|}{\textbf{Generation}} & \multicolumn{2}{c}{\textbf{Evaluation}} \\ \midrule
            & \textit{$\tau$} & \textit{$\tau$} & \textit{Agree.(\%)} \\ \midrule
            GPT-4o                 & {\ul 0.556}         & \textbf{0.944}      & 0.819  \\
            Gemini 1.5        & 0.444               & 0.889               & {\ul 0.846}  \\
            GPT-3.5                & 0.389               & 0.833               & 0.721  \\
            LLaMA 3.1 70B           & 0.5                 & 0.722               & \textbf{0.845}  \\
            Tulu 2.5 + PPO (70B RM) & 0.222               & 0.611               & \textbf{0.845}  \\
            Tulu 2.5 + PPO (13B RM) & 0.056               & 0.556               & 0.844  \\
            Mixtral  & \textbf{0.667}      & 0.556               & 0.845  \\
            Tulu 2.5 + DPO (13B)         & 0.511               & 0.809               & 0.684  \\ \bottomrule
        \end{tabular}
        \caption{Instruction-following}
        \label{tab:helpfulness_correlation}
    \end{minipage}%
    \hspace{0.02\textwidth} % 두 번째 표와 간격
    %================= 두 번째 subtable =================%
    \begin{minipage}[t]{0.32\textwidth} 
        \centering
        \begin{tabular}{@{}|c|cc@{}}
            \toprule
            \multicolumn{1}{|c|}{\textbf{Generation}} & \multicolumn{2}{c}{\textbf{Evaluation}} \\ \midrule
            \textit{$\tau$} & \textit{$\tau$} & \textit{Agree.(\%)} \\ \midrule
            {\ul 0.60}  & {\ul 0.778} & \textbf{0.654} \\
            {\ul 0.60}  & \textbf{0.822} & {\ul 0.61} \\
            0.467 & 0.378 & 0.551 \\
            0.60  & 0.689 & 0.605 \\
            0.067 & 0.200 & 0.520 \\
            0.333 & 0.378 & 0.526 \\
            \textbf{0.778} & -0.200 & 0.529 \\
            0.333 & 0.667 & 0.540 \\ \bottomrule
        \end{tabular}
        \caption{Document-based QA}
        \label{tab:helpfulness_correlation_2}
    \end{minipage}
    \end{tabular}
    \end{adjustbox}
    \caption{Model correlations (Kendall's $\tau$) with human values for helpful response generation tasks (SHP-2) and document-based QA tasks (WebGPT), and response-level agreement with human preferences.}
    \label{tab:non_summary_evaluation_correlation}
\end{table*}
{\subsection{Factor Correlations}}
{Figure~\ref{fig:corr_mat} presents the correlation matrix for the GPT-4o, Gemini-1.5, and Tulu 2.5 + PPO (13B RM) models across three tasks. The analysis focuses on the correlation between the distributions of feature scores for each feature within the samples generated by these models.}

{In summarization task, the patterns of feature correlation are generally consistent across the three models. Notably, there is a strong correlation between \{length and number of facts\} as well as \{number of facts and source coverage\}. These results are intuitive: the more factual content an answer includes, the longer the response tends to be, which in turn increases the likelihood of covering information from the source material.}

{In helpfulness task, 
All three models consistently exhibit a high correlation among \{length, number of facts, and helpfulness\}. This is expected, as longer responses are more likely to include a greater number of facts, which often translates into more helpful content. Interestingly, in the GPT-4o model specifically, there is a noticeable correlation between ``receptiveness'' and the set of factors \{helpfulness, number of facts, coherence, length\}. As detailed in Table~\ref{Appendix:full_result_shp2}, these are precisely the factors that GPT-4o tends to prioritize in this task. This pattern suggests that the GPT-4o model frequently considers these factors during response generation, resulting in a higher prevalence of these features in its outputs.}

{In the WebGPT task, there was a high correlation among \{length, number of facts, and helpfulness\}, similar to the helpfulness task. For GPT-4o and Tulu 2.5 + PPO (13B RM), the correlation between novel word and hallucination was high, which can be explained by the tendency to use novel words when hallucinating something.
}

\begin{figure*}[ht]
    \centering
    % --- Row 1: Summarization ---
    \begin{subfigure}[b]{\textwidth}
        \centering
        \begin{subfigure}[b]{0.32\textwidth}
            \centering
            \includegraphics[width=\textwidth]{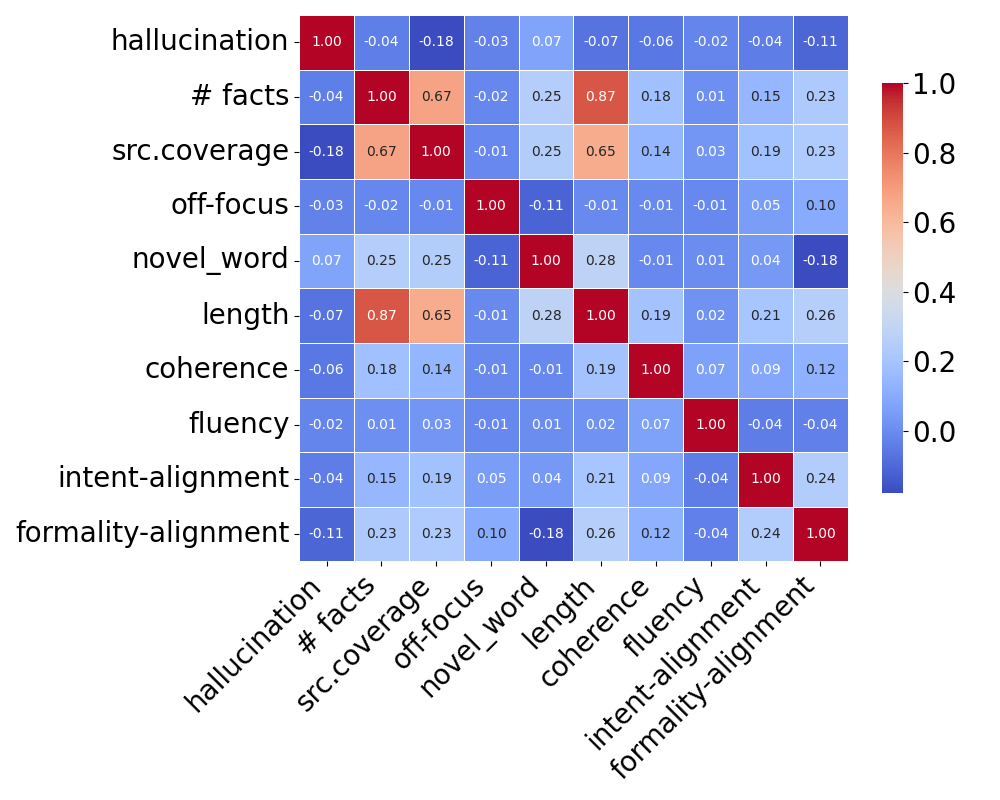}
            {\scriptsize GPT-4o}
        \end{subfigure}
        \hfill
        \begin{subfigure}[b]{0.32\textwidth}
            \centering
            \includegraphics[width=\textwidth]{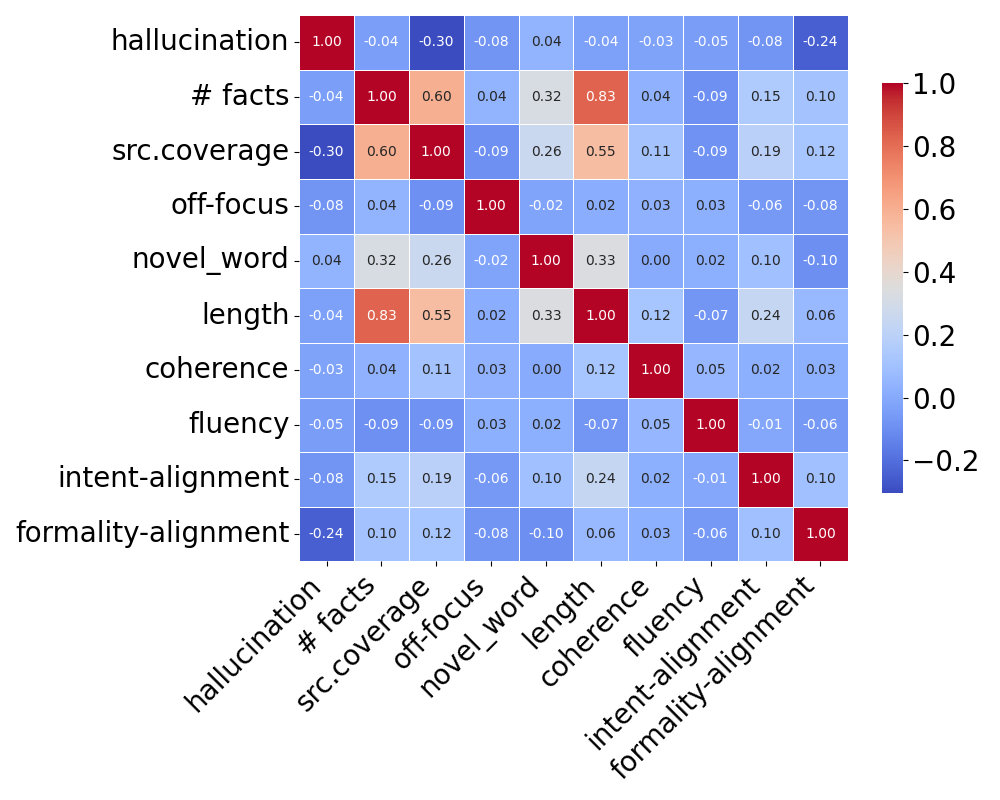}
            {\scriptsize Gemini-1.5}
        \end{subfigure}
        \hfill
        \begin{subfigure}[b]{0.32\textwidth}
            \centering
            \includegraphics[width=\textwidth]{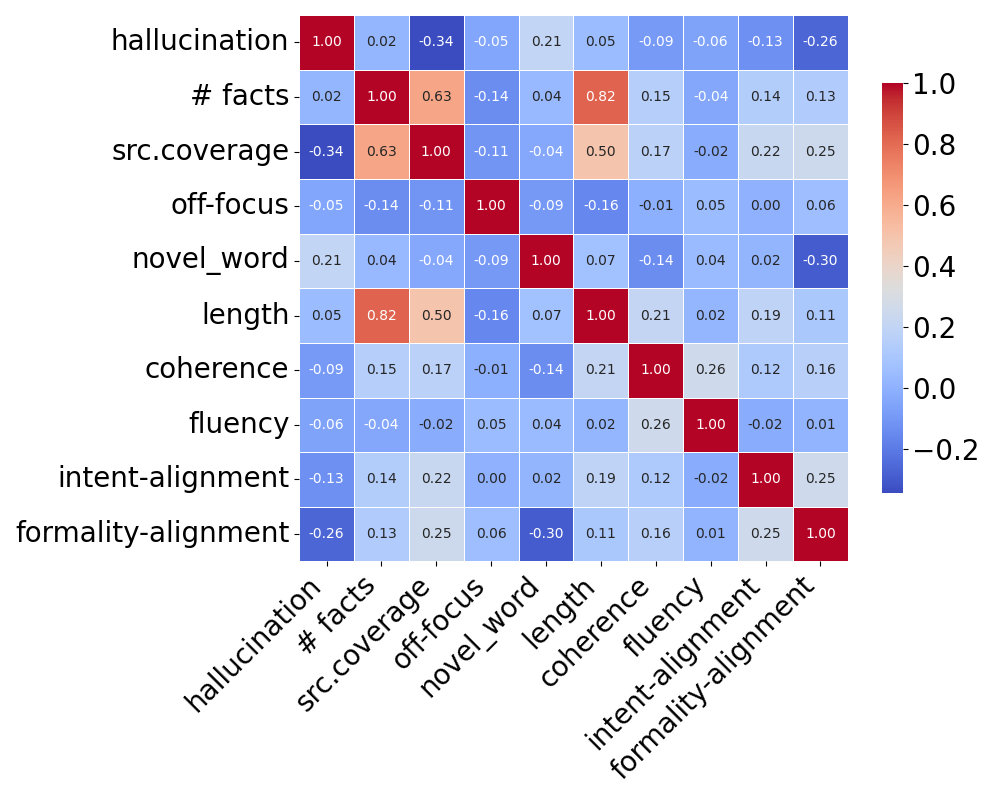}
            {\scriptsize Tulu 2.5 + PPO (13B RM)}
        \end{subfigure}
        \caption{Summarization}
    \end{subfigure}

    \vspace{1em} % Add space between rows

    % --- Row 2: SHP2 ---
    \begin{subfigure}[b]{\textwidth}
        \centering
        \begin{subfigure}[b]{0.32\textwidth}
            \centering
            \includegraphics[width=\textwidth]{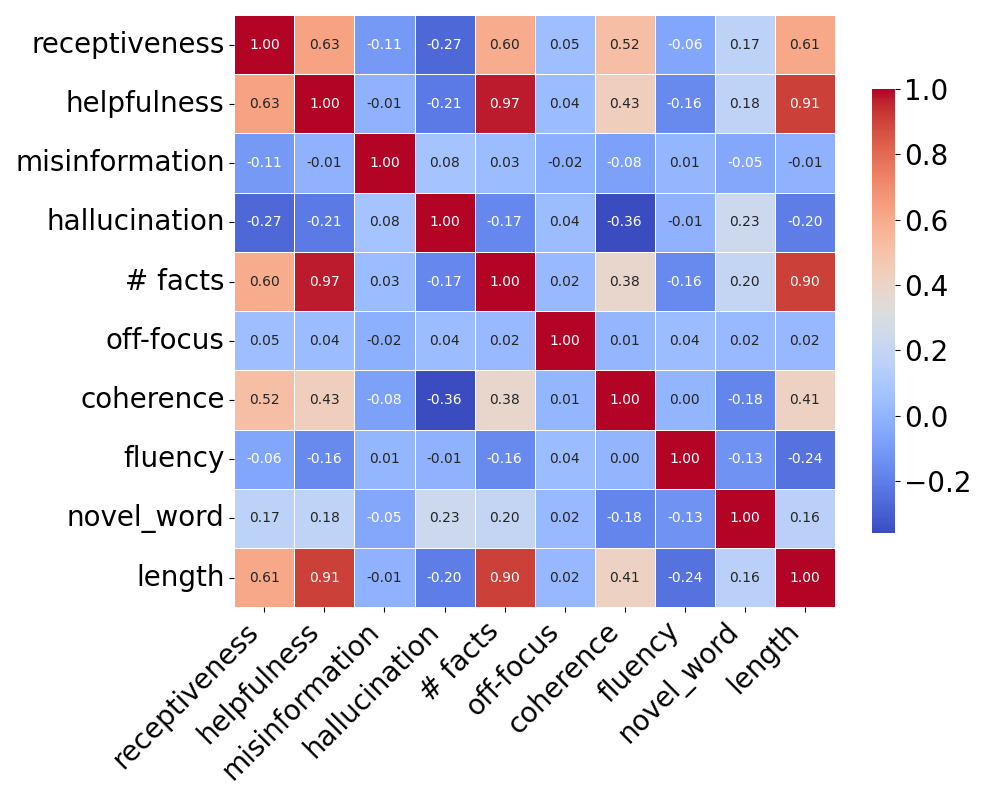}
            {\scriptsize GPT-4o}
        \end{subfigure}
        \hfill
        \begin{subfigure}[b]{0.32\textwidth}
            \centering
            \includegraphics[width=\textwidth]{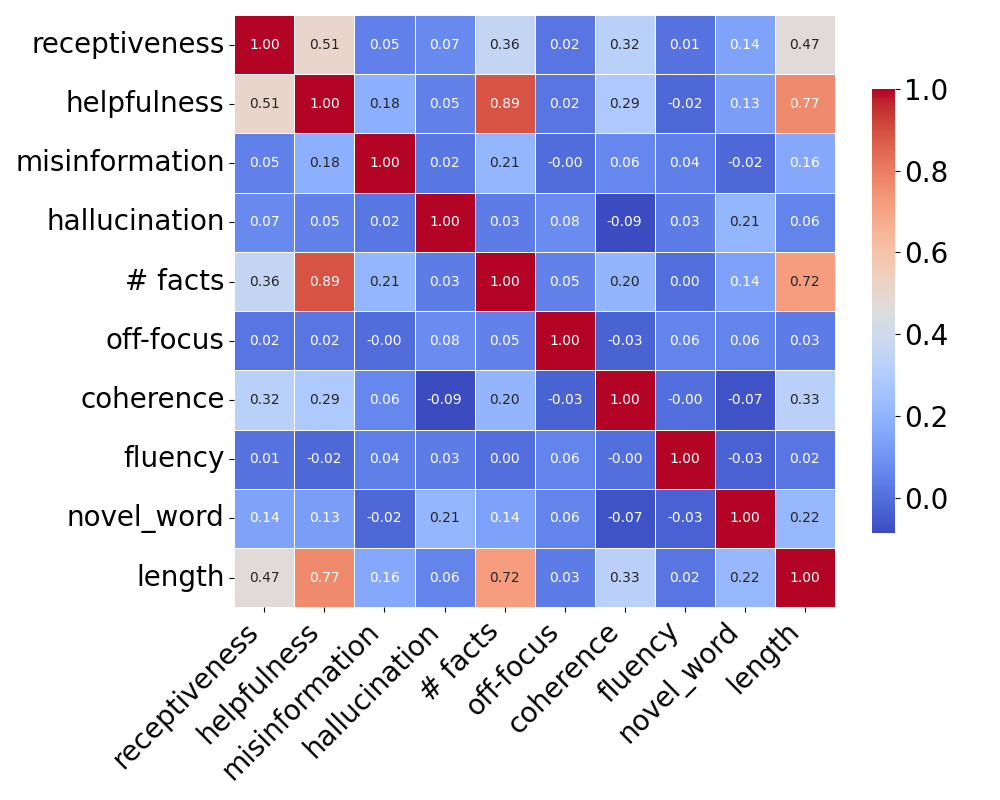}
            {\scriptsize Gemini-1.5}
        \end{subfigure}
        \hfill
        \begin{subfigure}[b]{0.32\textwidth}
            \centering
            \includegraphics[width=\textwidth]{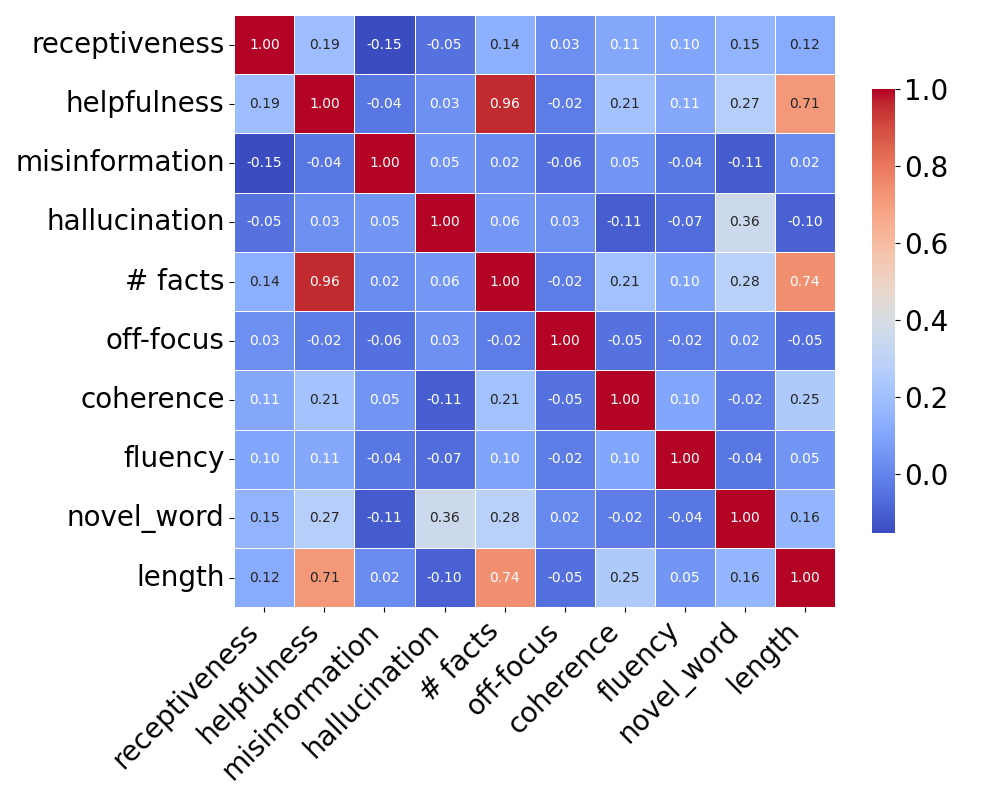}
            {\scriptsize Tulu 2.5 + PPO (13B RM)}
        \end{subfigure}
        \caption{Helpful Response Generation}
    \end{subfigure}

    \vspace{1em} % Add space between rows

    % --- Row 3: WebGPT ---
    \begin{subfigure}[b]{\textwidth}
        \centering
        \begin{subfigure}[b]{0.32\textwidth}
            \centering
            \includegraphics[width=\textwidth]{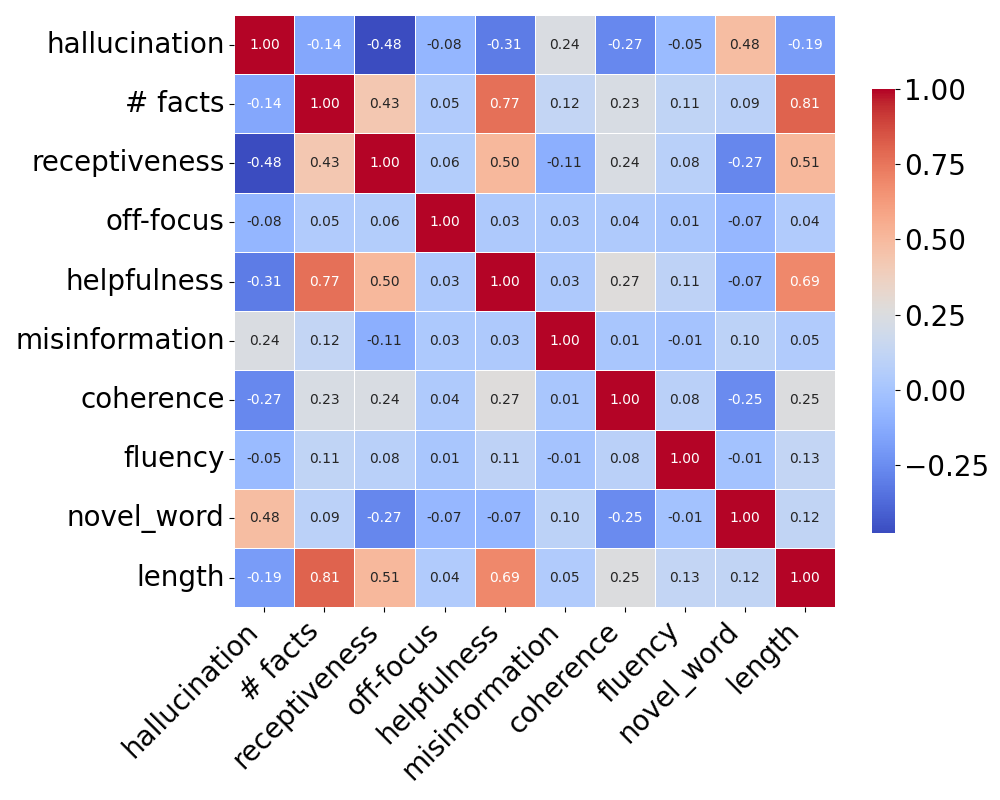}
            {\scriptsize GPT-4o}
        \end{subfigure}
        \hfill
        \begin{subfigure}[b]{0.32\textwidth}
            \centering
            \includegraphics[width=\textwidth]{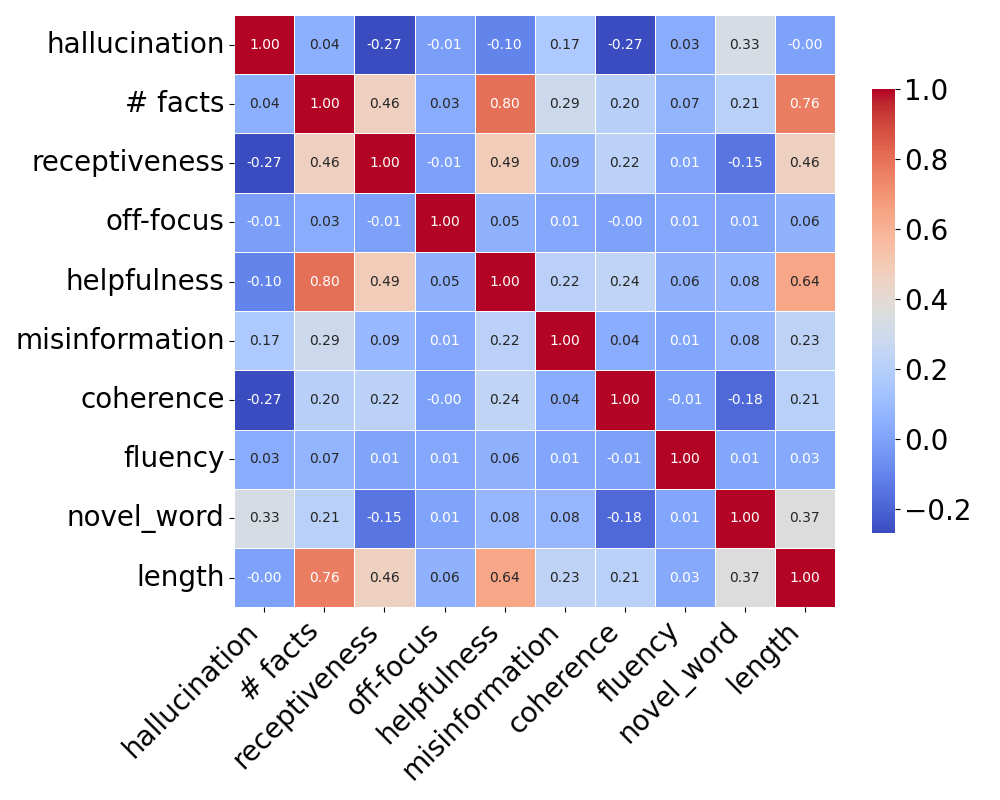}
            {\scriptsize Gemini-1.5}
        \end{subfigure}
        \hfill
        \begin{subfigure}[b]{0.32\textwidth}
            \centering
            \includegraphics[width=\textwidth]{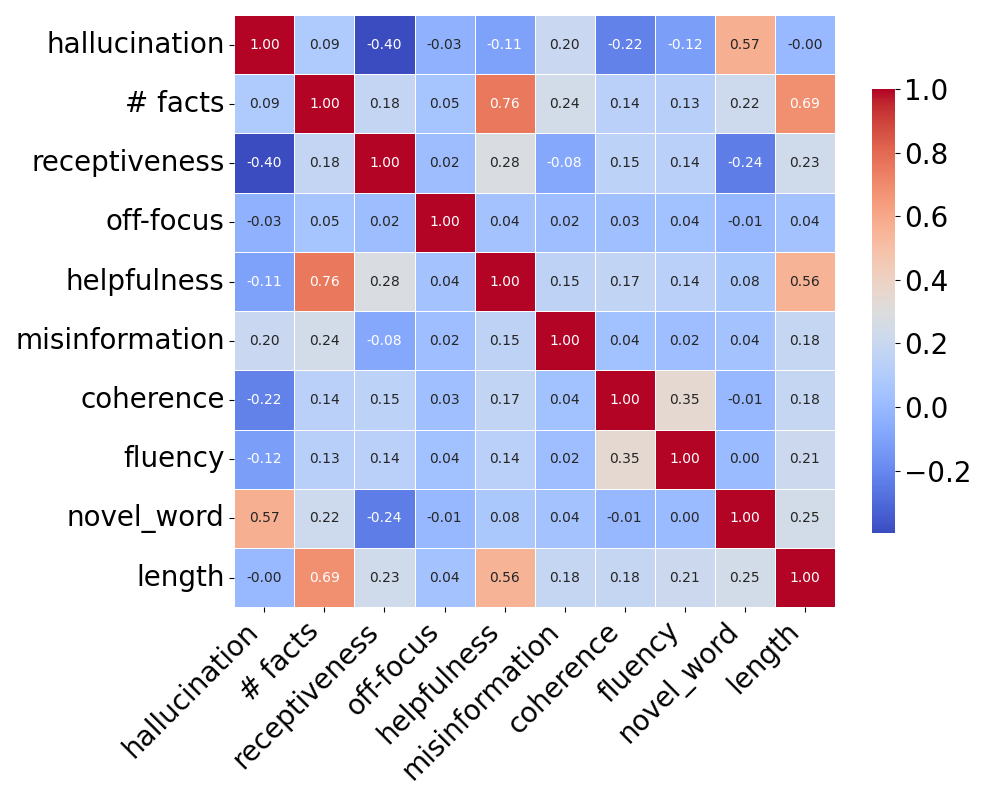}
            {\scriptsize Tulu 2.5 + PPO (13B RM)}
        \end{subfigure}
        \caption{Document-based QA}
    \end{subfigure}

    \caption{Correlation matrices for various models across tasks.}
    \label{fig:corr_mat}
\end{figure*}

% Figure~\ref{fig:corr_mat} shows the correlation matrix of GPT-4o, Gemini-1.5, and Tulu 2.5 + PPO (13B RM) models across three tasks. We analyze the correlation between the distributions of feature scores for each feature in the samples generated by the model. 
% summarization
% - The tendency of feature correlation was similar in the three models.
% - Representatively, there is a high correlation between {length and number of facts} and {number of facts and source coverage} independently. It is intuitive that the larger the number of facts an answer contains, the longer the answer and the higher the likelihood of containing information from the source.
% ### 2. SHP2
% - All three models consistently show a high correlation between every pair among {length, number of facts, and helpfulness}. This result is expected because, in addition to the explanation between {length and number of facts} in above, longer responses are more likely to include helpful advice or information.
% - Only in GPT-4o, there is a noticeable trend of high correlation between "receptiveness" and {helpfulness, number of facts, coherence, length}. Interestingly, as shown in Table 6 in Appendix F, these four factors are the ones that GPT-4o tends to prioritize in this task.
% - As the gpt-4o model frequently considers the corresponding factors during generation, it seems to be the result of the frequent appearance of the corresponding feature in the output.

\section{Generalizability of Our Results}
\label{sec:appendix_generalizability}
\begin{table*}[!htpb]
\centering
\small
\begin{tabular}{@{}c|l|lllll@{}}
\toprule
Task                               & \multicolumn{1}{c|}{Model} & \multicolumn{1}{c}{Score 1} & \multicolumn{1}{c}{Score 2} & \multicolumn{1}{c}{Score 3} & \multicolumn{1}{c}{Score 4} & \multicolumn{1}{c}{Score 5} \\ \midrule
\multirow{3}{*}{Summarization}           & GPT-4o                     & 0.791                       & 0.823                       & 0.856                       & 0.886                       & \textbf{0.901}              \\
                                   & Tulu 2.5 + PPO (70B RM)    & 0.831                       & 0.852                       & 0.850                       & 0.856                       & \textbf{0.863}              \\
                                   & LLaMA 3.1 70B              & 0.711                       & 0.792                       & 0.828                       & 0.849                       & \textbf{0.854}              \\ \midrule
\multirow{3}{*}{Helpful Response Generation}       & GPT-4o                     & 0.532                       & 0.604                       & 0.620                       & 0.637                       & \textbf{0.685}              \\
                                   & Tulu 2.5 + PPO (70B RM)    & 0.435                       & 0.492                       & 0.581                       & 0.641                       & \textbf{0.679}              \\
                                   & LLaMA 3.1 70B              & 0.463                       & 0.516                       & 0.628                       & 0.662                       & \textbf{0.690}              \\ \midrule
\multirow{3}{*}{Document-based QA} & GPT-4o                     & 0.528                       & 0.599                       & 0.625                       & 0.657                       & \textbf{0.697}              \\
                                   & Tulu 2.5 + PPO (70B RM)    & 0.513                       & 0.572                       & 0.631                       & 0.691                       & \textbf{0.738}              \\
                                   & LLaMA 3.1 70B              & 0.532                       & 0.570                       & 0.644                       & 0.706                       & \textbf{0.765}              \\ \bottomrule
\end{tabular}

\caption{Comparison of similarity between directly generated responses and score-based responses for summarization, helpful response generation, and document-based QA tasks.}
\label{tab:response_score_similarity}
\end{table*}

Our research deviates from the typical language model setup by using a 1-5 scoring system for response generation. To assess the validity of our approach, we compare responses generated through direct generation (without scoring) with those across the score range {through all summary, helpfulness, and document-based QA tasks}. {In every task,} we found that score 5 consistently aligns best with direct generation responses, based on the fine-grained factors we use, in models like GPT-4o, Tulu 2.5 + PPO (70B RM), and LLaMA 3.1 70B (see Table~\ref{tab:response_score_similarity} in the Appendix~\ref{sec:appendix_generalizability}). This suggests that our scoring framework, specifically score 5, captures the essence of unconstrained language model outputs, implying the potential generalizability of our findings to general settings.

We conduct experiments by prompting the model to generate responses with scores ranging from 1 to 5. This setup allows us to verify whether the results can generalize to a typical scenario where the model generates responses directly. We compare the model's direct responses and the score-based responses for the summarization task on Reddit TL;DR using outputs from GPT-4o, Tulu 2.5 + PPO (70B RM), and LLaMA 3.1 70B.

Since the value ranges differ across features, we scale the data using min-max scaling before calculating cosine similarity. The results in Table~\ref{tab:response_score_similarity} indicate that the model's direct responses are most similar to those with a score of 5, all showing a high similarity of over 0.85. Overall, as the scores decrease, the similarity also declines.

This finding suggests that the model's direct responses align closely with its best-generated responses. Additionally, the lower the score, the less similarity there is to the direct responses, indicating that our score-based responses align well with the model's outputs. Thus, we demonstrate that our findings can generalize to typical settings where responses are generated directly by the model.

\section{Use of AI Assistant}
We used ChatGPT web assistant (ChatGPT Pro)\thinspace\footnote{\url{https://chatgpt.com/}} and Gemini web application (2.0 Flash)\thinspace\footnote{\url{https://gemini.google.com/}} to refine the writing of the manuscript.

\end{document}